\documentclass[10pt,twocolumn,letterpaper]{article}

\usepackage[pagenumbers]{cvpr} 

\usepackage[utf8]{inputenc} 
\usepackage[T1]{fontenc}    

\usepackage{url}            
\usepackage{booktabs}       
\usepackage{amsfonts}       
\usepackage{nicefrac}       
\usepackage{microtype}      
\usepackage{xcolor}         
\usepackage{subfiles}
\usepackage{bm}
\usepackage{multirow}
\usepackage[accsupp]{axessibility} 
\usepackage{graphicx}
\usepackage{wrapfig}
\usepackage[font=small,labelfont=bf]{caption}
\usepackage{amsmath}
\usepackage{pifont}
\usepackage{colortbl}
\usepackage{tabularx}
\usepackage{arydshln}
\usepackage{makecell}
\usepackage{tikz}

\usepackage{siunitx}
\include{manuscripts/tables}
\include{manuscripts/figures}
\newcommand{\Paragraph}[1]{\vspace{1.5mm}\noindent\textbf{#1.}\hspace{0.5mm}}

\definecolor{mygreen}{rgb}{0.0, 0.8, 0.6}

\newcommand{\token}{
    \tikz\draw[thick] (0,0) rectangle (0.2,0.2) node[pos=.5, text=white, inner sep=0] {\tiny 1};
}
\newcommand{\maskednumtoken}[1]{
    \tikz\filldraw[thick, black] (0,0) rectangle (0.2,0.2) node[pos=.5, text=white, inner sep=0] {\tiny #1};
} 

\definecolor{cvprblue}{rgb}{0.21,0.49,0.74}
\usepackage[pagebackref,breaklinks,colorlinks,citecolor=cvprblue]{hyperref}

\def\paperID{6668} 
\def\confName{CVPR}
\def\confYear{2025}

\title{SapiensID: Foundation for Human Recognition}

\author{Minchul Kim$^1$, \quad Dingqiang Ye$^1$, \quad 
Yiyang Su$^1$, \quad Feng Liu$^2$,
\quad Xiaoming Liu$^1$ \\
\textsuperscript{1} Department of Computer Science and Engineering, Michigan State University\\
\textsuperscript{2} Department of Computer Science, Drexel University\\
{\tt\small $^1$\{kimminc2, yedingqi, suyiyan1, liuxm\}@msu.edu,}
{\tt\small $^2$fl397@drexel.edu}
}

\begin{document}

\twocolumn[{%
\renewcommand\twocolumn[1][]{#1}%
\maketitle
\begin{center}
    \centering
    \vspace{-1mm}
    \captionsetup{type=figure}
    \includegraphics[width=\linewidth]{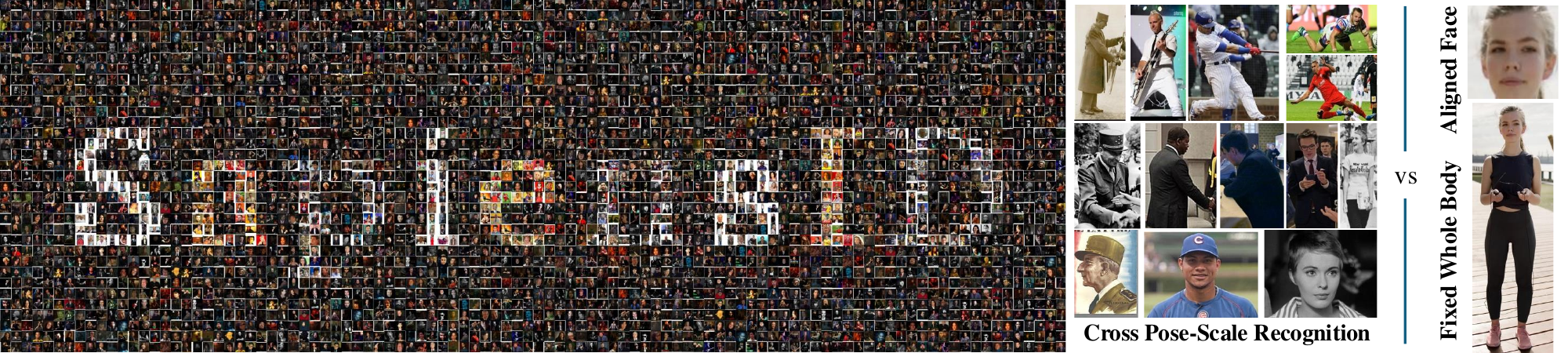}
    \captionof{figure}{SapiensID is a human recognition model trained on a large-scale dataset of human images featuring varied poses and visible body parts. For the first time, a {\it single} model performs effectively across diverse face and body benchmarks~\cite{lfw,calfw,shu2021large,yang2019person}. This marks a significant improvement over previous body recognition models, which were often limited to one specific camera setup or image alignments for one model, with worse performance in in-the-wild scenarios. Additionally, we introduce a large-scale, cross-pose and cross-scale training and evaluation set designed to facilitate further research in this area. --- The name SapiensID pertains to the ability to recognize humans.}
    \label{fig:one}
\end{center}%
}]

\maketitle

\begin{abstract}
 Existing human recognition systems often rely on separate, specialized models for face and body analysis, limiting their effectiveness in real-world scenarios where pose, visibility, and context vary widely. This paper introduces SapiensID, a unified model that bridges this gap, achieving robust performance across diverse settings. SapiensID introduces (i) Retina Patch (RP), a dynamic patch generation scheme that adapts to subject scale and ensures consistent tokenization of regions of interest, (ii) a masked recognition model (MRM) that learns from variable token length, and (iii) Semantic Attention Head (SAH), an module that learns pose-invariant representations by pooling features around key body parts. To facilitate training, we introduce WebBody4M, a large-scale dataset capturing diverse poses and scale variations. Extensive experiments demonstrate that SapiensID achieves state-of-the-art results on various body ReID benchmarks, outperforming specialized models in both short-term and long-term scenarios while remaining competitive with dedicated face recognition systems. Furthermore, SapiensID establishes a strong baseline for the newly introduced challenge of Cross Pose-Scale ReID, demonstrating its ability to generalize to complex, real-world conditions. \href{https://github.com/mk-minchul/sapiensid}{Project Link}
\end{abstract}

\section{Introduction}

Human recognition has traditionally been approached through domain-specific models focused exclusively on either face~\cite{deng2019arcface,wang2018cosface,kim2022caface,kim2022adaface,wang2017normface,liu2017sphereface,kim2020broadface,huang2020curricularface,kim2023dcface,huang2020improving,yin2020fan,wheeler2007multi} or body~\cite{yang2019person, li2021learning,gu2022clothes,jin2022cloth,liu2024distilling,3DInvarReID} recognition (or ReID). Each of these modalities relies heavily on specific dataset alignments, where face recognition models are optimized for tightly cropped, aligned facial images~\cite{deng2020retinaface,msceleb,zhu2021webface260m,insightface}, and body recognition models are designed to process full-body images of standing individuals ~\cite{yang2019person,shu2021large,market,msmt17}.

\begin{figure}[t!]
    \centering
    \includegraphics[width=0.9\linewidth]{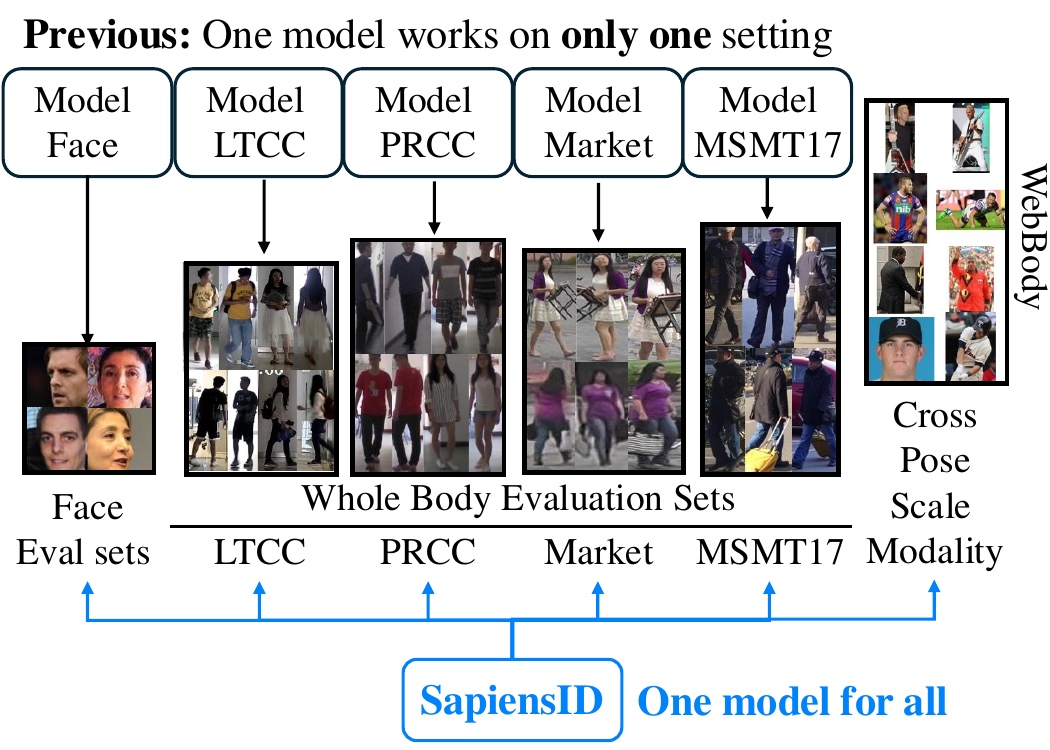}
     \vspace{-3mm}
    \caption{Conventionally, face and body recognition were handled independently. Also body models are trained on one specific dataset without the ability to generalize to other datasets. SapiensID model for the first time generalizes across modalities and different body poses and camera settings. \vspace{-3mm}}
    \label{fig:fig2}
\end{figure}

Despite advances in face and body recognition, no single model has yet effectively managed to handle a diverse range of poses and visible area simultaneously. 
However, in real-world settings, human recognition often requires harnessing the full spectrum of available clues, integrating both face and body information. Typically, multiple models are fused at the feature or score level~\cite{he2010performance,liu2024farsight} to mitigate this issue. In other words, no single model can handle both face and body images as robustly as modality-specific models. Therefore, a unified model would mark a significant advance in human recognition, allowing reliable identification across varied poses and scales of body parts. 
As in Fig.~\ref{fig:fig2}, current models relies heavily on in-domain datasets, fail to generalize effectively to other datasets. 

Addressing this gap is important for several reasons.
In real-world applications, human recognition systems should operate across a variety of poses (sitting vs standing) and visible contextual areas (upper torso vs whole body)~\cite{yao2010modeling}. For instance, IJB-S~\cite{ijbs} contains face gallery images and whole body probe videos. Furthermore, a unified model simplifies model deployment and usage for downstream tasks by eliminating the need for preprocessing steps such as face alignment~\cite{deng2020retinaface} or dependency on camera setups~\cite{shu2021large,yang2019person}.

However, addressing this problem is not trivial. First, it requires a large-scale labeled human image dataset that captures a wide range of poses and visibility variations. Secondly, even with such a dataset, the model must be capable of managing the substantial variability in scale and pose that human images naturally show. As in Fig.~\ref{fig:one}, close-up portraits show a large face, while full-body shots display it much smaller.
Modality-specific models have eliminated the scale inconsistency problem with some form of pre-alignment stage. For instance, body recognition models assume consistent camera setup~\cite{shu2021large,yang2019person} and face recognition models assume the images are aligned with 5 facial landmarks to a canonical position~\cite{zhang2016joint,deng2020retinaface}. 
Such transformations of input reduce irrelevant variability in recognizing a person, making training easier. However, models fail to generalize when the preprocessing step fails~\cite{kprpe}. 

To this end, we propose \textbf{SapiensID}, one model capable of handling the complexities of human recognition in diverse settings. Our contributions are

\begin{itemize}
    \item \textbf{Model Innovations:} We introduce three major improvements over conventional specialized recognition models:
    \begin{enumerate}
        \item \textbf{Retina Patch} addresses scale variations often encountered in human images by dynamically allocating more patches to important regions.
        \item \textbf{Masked Recognition Model} reduces the number of tokens, achieving $8\times$ speed up in ViT during training.
        \item \textbf{Semantic Attention Head} addresses pose variations by learning to pool features around keypoints. 
    \end{enumerate}
    
    \item \textbf{Data Contribution:} To aid the development and evaluation of \textbf{SapiensID}, we release \textbf{WebBody4M} (Fig.~\ref{fig:one}), a large-scale dataset specifically designed for comprehensive human recognition across different poses and scales.

    \item \textbf{Performance:} SapiensID achieves state-of-the-art results across multiple ReID benchmarks and establishes a strong baseline for the novel Cross Pose-Scale ReID task.
\end{itemize}

Our approach is a paradigm shift human recognition, laying the groundwork for research that bridges the gap between specialized models and holistic recognition systems.

\section{Related Works}

\subsection{Face Recognition}

Face Recognition (FR) matches query images to an enrolled identity database. 
State-of-the-art (SoTA) FR models are trained on large-scale  datasets~\cite{zhu2021webface260m, msceleb, deng2019arcface} with margin-based softmax losses~\cite{wang2018cosface, deng2019arcface, liu2017sphereface, huang2020curricularface, kim2022adaface}. 
FR performance is evaluated on a set of benchmarks, {\it e.g.} LFW~\cite{lfw}, CFP-FP~\cite{cfpfp}, CPLFW~\cite{cplfw}, AgeDB~\cite{agedb},  CALFW~\cite{calfw}, and IJB-B,C~\cite{ijbb,ijbc}. 
They are designed to assess the model's robustness to factors such as pose variations and age differences.  
Models trained on large datasets,  {\it e.g.} WebFace260M, achieve over $97\%$ verification accuracy  on these benchmarks~\cite{kim2022adaface}. 
FR in low-quality imagery is substantially harder and TinyFace~\cite{tinyface} and IJB-S~\cite{ijbs} are popular benchmarks.

Face recognition is often accompanied by facial landmark prediction~\cite{zhang2016joint, bulat2017far,tai2019towards,luvli-face-alignment-estimating-landmarks-location-uncertainty-and-visibility-likelihood} so that input faces are aligned and tightly cropped around the facial region. 
However, when alignment fails, FR models perform poorly~\cite{kprpe}. 
Eliminating alignment would not only simplify the pipeline but also enhance robustness in conditions where alignments are prone to fail. 
We propose an {\it alignment-free} paradigm capable of handling any human image with or without a visible face.

\subsection{Body Recognition}
Body recognition, {\it a.k.a.}~Person Re-identification (ReID), seeks to identify individuals across different times, locations, or camera settings. 
Prior works~\cite{wang2018transferable,li2018unsupervised, li2019unsupervised, lin2019bottom,yu2019unsupervised, zhai2020ad,ge2020mutual,ge2020self,market} focus on short-term scenarios where subjects generally end up with the same attire. 
Removing this assumption has led to long-term, cloth-changing 
ReID~\cite{yang2019person, li2021learning,chen2021learning,hong2021fine,gu2022clothes,jin2022cloth,wan2020person,yu2020cocas,su2024open}, 
on datasets like PRCC~\cite{yang2019person}, LTCC~\cite{shu2021large}, CCDA~\cite{3DInvarReID} and CelebReID~\cite{huang2019celebrities, huang2019beyond}. 

All of these datasets are composed primarily of whole-body images, where the subjects are fully visible from head to toe, with poses generally limited to walking or standing. 
While this format has been valuable in the development of person ReID models for controlled environments, it lacks the scale and visibility variety often encountered in real-world applications. To address these limitations, we propose a model capable of handling diverse and complex poses and visible areas. 
Further, to facilitate the training and evaluation of these models, we introduce a new large-scale, labeled dataset that significantly broadens pose-scale diversity.

\subsection{Patch Generation for Vision Transformers}

In Vision Transformer (ViT)~\cite{vit}, an image is divided into patches, with each transformed into a token via linear projection. This patch-based approach transforms images to an unordered set of tokens for sequence-to-sequence modeling~\cite{vaswani2017attention}, processing images in a scalable and flexible way in downstream tasks. Typically, patches are created by dividing an image into a grid with a specific number of patches.

Several works explore how the patchifying process helps ViT capture multi-scale objects in images~\cite{dynavit}. 
For instance, \cite{navit} predefines patch counts without resizing the input,  retaining the image’s aspect ratio and scale. 
 \cite{flexivit} randomizes patch sizes in training for generalization across image scales,  enhancing efficiency while sometimes reducing accuracy. 
Importantly, the representation quality of specific regions, such as face or hand, depends on \textbf{the number of tokens} allocated to those areas. 
A smaller face within a constant patch size, for example, generates fewer tokens and thus captures less detail than a larger face. 
To address this, we propose to maintain a consistent number of tokens for regions of interest while ensuring full, non-overlapping coverage across the image in line with grid-based tokenization principles.

\section{Proposed Method}
A human recognition model is formulated as a metric learning task such that images of the same subject are closer in feature space than those of different subjects, satisfying 
\begin{equation}
    d(\mathbf{f}_A^i, \mathbf{f}_A^j) < d(\mathbf{f}_A^i, \mathbf{f}_B^k),
\end{equation}  
where \( \mathbf{f}_A^i \) and \( \mathbf{f}_A^j \) denote the feature vectors of two different images \( i \) and \( j \) of the same subject \( A \), while \( \mathbf{f}_B^k \) represents the feature vector of an image of a different subject \( B \). Notably, the subjects \( A \) and \( B \) are not observed during training. Following established research on margin-based techniques for enhancing intra-class compactness in the feature space~\cite{deng2019arcface, kim2022adaface, wang2018cosface, liu2017sphereface, meng2021magface}, we utilize a margin-based softmax loss~\cite{kim2022adaface} to train our model on a labeled dataset. We collect a large-scale web-collected human image training dataset which will be discussed in Sec.~\ref{sec:webbody}.  

The key challenge that sets this apart from prior work on a separate face~\cite{deng2019arcface,  liu2017sphereface} or body~\cite{yang2019person, li2021learning} recognition task is that the input image can be highly varying in 1) scale and 2) body pose. 
To tackle these challenges, we propose a new architecture, which will be discussed in the subsections.

\begin{figure}[t]
    \centering
    \includegraphics[width=0.8\linewidth]{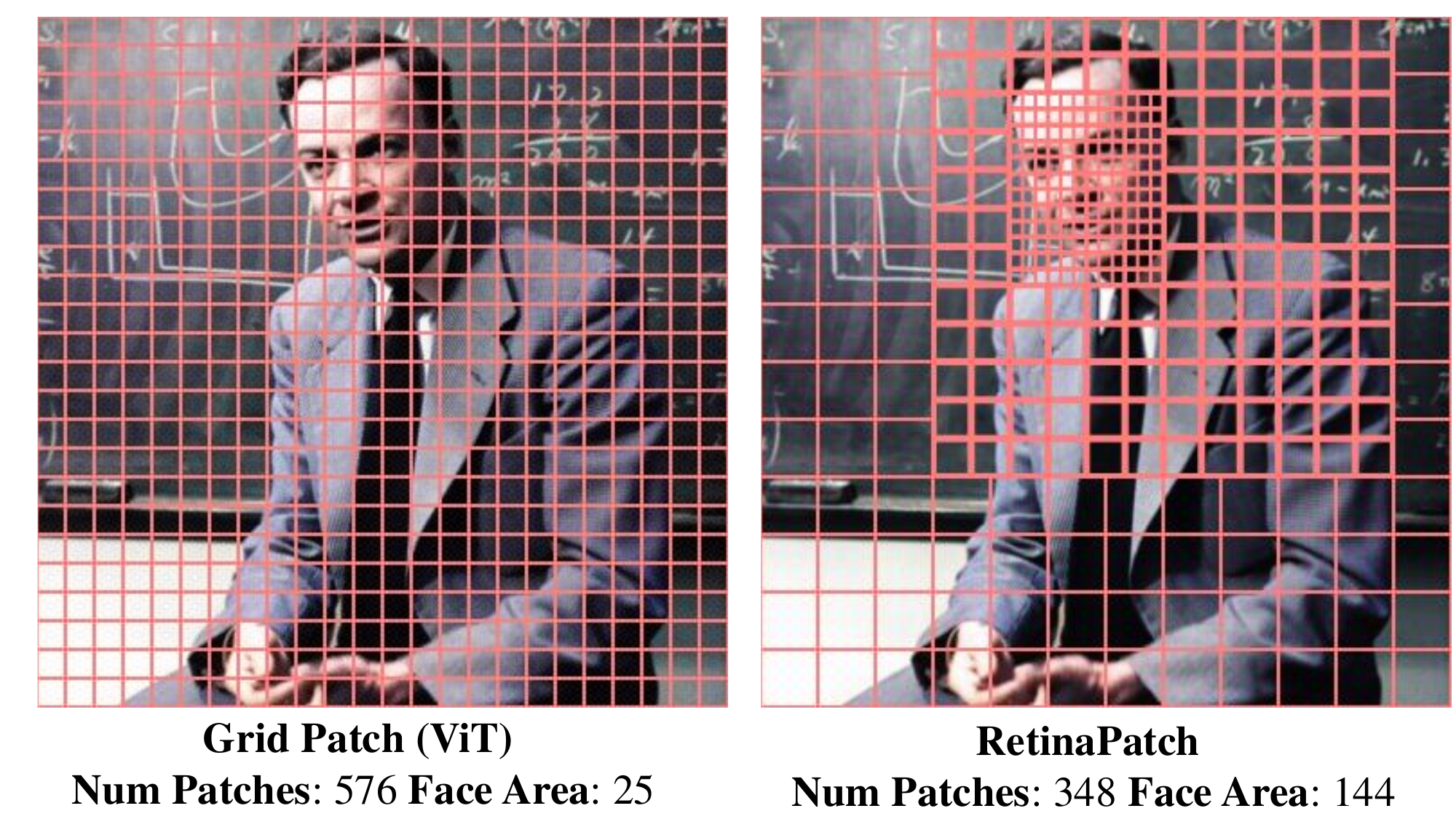}
    \vspace{-2mm}
        \caption{Comparison between the standard grid patch scheme of Vision Transformers (ViT) and our Retina Patch. While maintaining the same or lower computational budget (number of tokens), Retina Patch dynamically allocates more patches to critical regions (e.g., face and upper torso) in an image. This allocation enhances the model's ability to capture fine-grained details in important regions, and to handle varying scales more effectively than fixed grid patch.}

    \label{fig:retina_patch}
    \vspace{-3mm}
\end{figure}

\subsection{Retina Patch (RP)}

To address the issue of varying scale in human images, we propose a novel \textbf{Retina Patch} mechanism inspired by the human eye’s ability to adapt focus dynamically to regions of interest (ROIs) within a scene. In natural images, subjects can appear in diverse poses and with varying visibility of the face and body, leading to substantial differences in scale across regions. For instance, in a full-body image, a face may be a small portion, whereas in a close-up, it dominates. To account for these variations, our Retina Patch dynamically assigns more patches to critical regions within the image.

Assume we have an input image $i$ and a set of image-dependent regions of interest, \( \{ \text{ROI}_{r}^i \mid r = 0, 1, \ldots, R \} \), each defined by a bounding box. 
There are \( R \) ROIs per image. 
Details on how ROIs are computed will be discussed later. 
We also let \( \text{ROI}_{0}^i \) be the whole image. 
For each \( \text{ROI}_{r}^i \), we set a specific number of patches \( m_{r} \) and an order \( z_{r} \), both controlling how many patches can come from each \( \text{ROI}_{r}^i \).

To obtain patches, we may perform a grid patching operation on each ROI independently. 
However, this would naturally result in overlapping patches with redundant feature extraction. 
Our aim is to cover the whole image with patches \textit{without any overlap}. 
To avoid redundancy, overlapping patches between regions with a lower order ({\it e.g.}, order \( z = 1 \)) and those with a higher order ({\it e.g.}, order \( z = 2 \)) are excluded from the patch set of the low-order regions. This selective inclusion process ensures that each patch belongs uniquely to the ROI with the highest priority, as indicated by the order. Specifically,
\begin{equation}
\text{P}^{i} = \bigcup_{r_1=0}^{R} \left( \text{P}_{\text{ROI}_{r_1}}^{i} - \bigcup_{r_2=r_1+1}^{R} \text{P}_{\text{ROI}_{r_2}}^{i} \right),
\end{equation}
where \( \text{P}_{\text{ROI}_{r}^i} \) represents the set of patches for region \( \text{ROI}_{r} \) of image \( i \), and \( r \) denotes the index of each ROI, ordered by their respective priorities for patch inclusion.

\begin{figure}[t]
    \centering
    \includegraphics[width=1.0\linewidth]{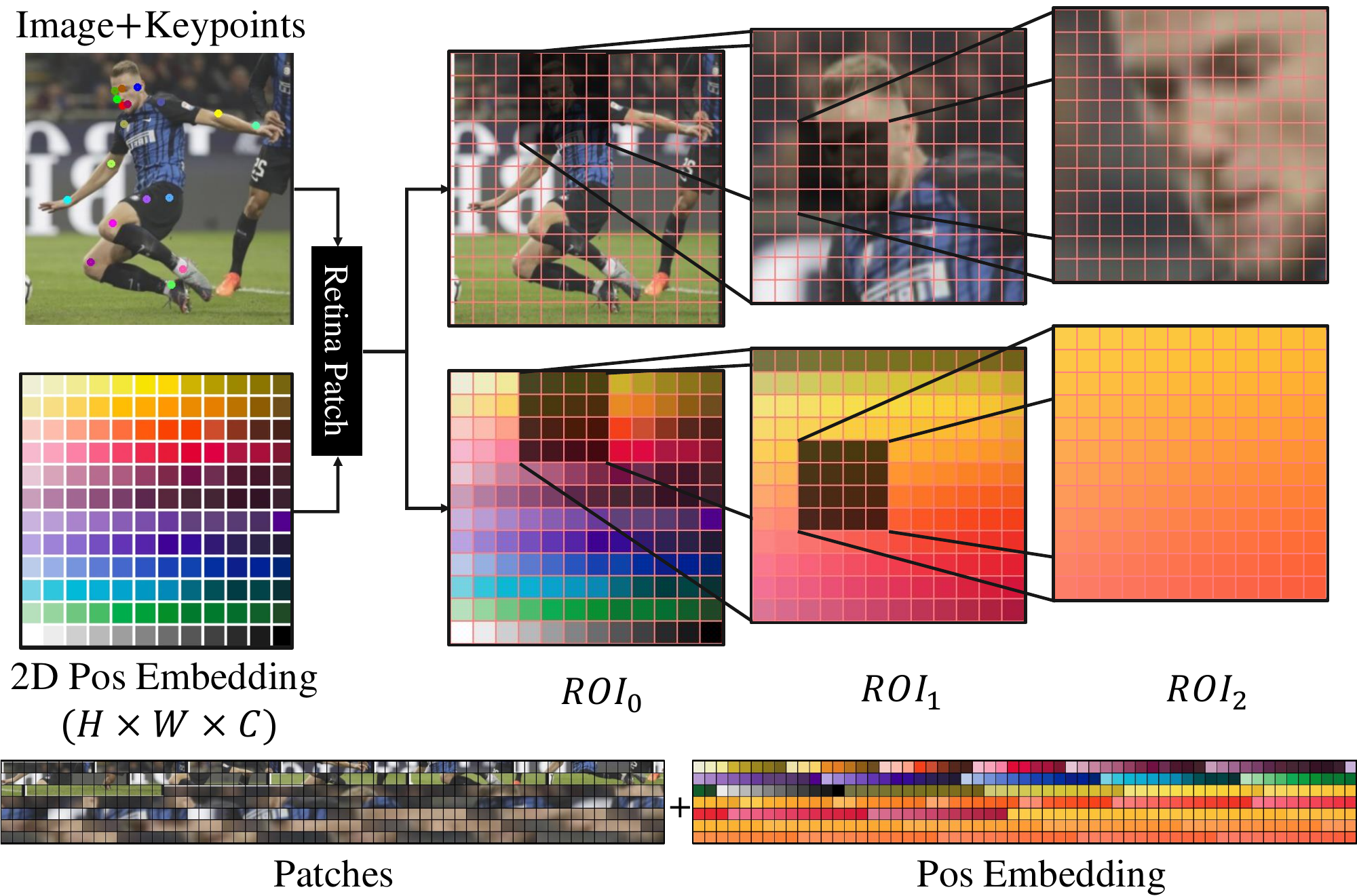}
    \vspace{-4mm}
    \caption{Illustration of Retina Patch and Position Encoding computation. \textbf{Top:} It shows three different ROIs generating patches at various scales (e.g., full image, upper torso, face). It also shows the corresponding position encodings sampled from the same spatial locations as the patches, allowing ViT to infer spatial context and understand where each patch originated within the image. \textbf{Bottom:} patches and position embedding created by Retina Patch. \vspace{-3mm}}
    \label{fig:rspe}
    
\end{figure}
This approach allows us to dynamically allocate critical regions with more patches while ensuring that the entire image is represented by patches without repetition. Also, the scale inconsistency is mitigated as long as the ROIs are semantically defined (\textit{e.g.}, \textit{face, upper torso}). The number of patches within each ROI is kept consistent across images, ensuring that each patch covers a similar scale within its designated ROI. Fig.~\ref{fig:retina_patch} uses an example to  compare the vanilla grid patch of ViT with our proposed Retina Patch.

\Paragraph{Computing ROI}
Retina Patch is a generic algorithm that can work for any class of images by designing ROIs for the particular domain. In this paper, for recognizing a subject from a human image, we set the ROIs in 3 parts: 1) whole image, 2) upper torso and 3) face. The upper torso and face ROIs are computed using the off-the-shelf body keypoint detector~\cite{openpose}. Details on transforming the keypoints into a bounding box can be found in Supp.

\Paragraph{Tokenization} 
The input to ViT's transformer block is a set of tokens or feature vectors. Since each patch's size is dependent on both the ROI size and the number of patches \( m_r \), the size of each patch may not be the same across ROIs. We simply resize all patches to be the size of patches from the whole image \( \text{ROI}_{0}^i \). We then use a linear layer to map each patch to the desired dimension, as in ViT.

\Paragraph{Position Embedding}
Since Transformer operates on sets of tokens without inherent order, Position Embedding (PE) is crucial for informing ViT of the spatial origin of each patch within the original image. For tokens of Retina Patch, we cannot use a traditional PE as the patch's source location is dynamic. 
Thus, we propose a Region-Sampled PE.

Let \( \text{PE} \in \mathbb{R}^{C \times H \times W} \) be the fixed 2D sin-cosine position embedding~\cite{chen2021empirical,beyer2022better} for the whole image. Given a normalized region of interest \( \text{ROI}_{r}^i = (x_r^i, y_r^i, h_r^i, w_r^i) \) with values between 0 and 1, we define a sampling grid \( \text{Grid}_{\text{ROI}_{r}^i} \) over the region \( [x_r^i, x_r^i + w_r^i] \) and \( [y_r^i, y_r^i + h_r^i] \) within the position embedding \( \text{PE} \). Let \( (h_r', w_r') \) be the target output shape for \( \text{PE}_{\text{ROI}_{r}^i} \), such that \( h_r' \cdot w_r' = m_r \), the desired number of patches for \( \text{ROI}_{r}^i \). The Region Sampled PE, \( \text{PE}_{\text{ROI}_{r}^i} \) is then obtained by bilinearly interpolating \( \text{PE} \) at the points in \( \text{Grid}_{\text{ROI}_{r}^i} \) to match the shape \( (h_r', w_r') \):
\begin{equation}
    \text{PE}_{\text{ROI}_{r}^i} = \text{GridSample}(\text{PE}, \text{Grid}_{\text{ROI}_{r}^i}, (h_r', w_r')) + v_r.
\end{equation}
We add a leanable parameter $v_r\in \mathbb{R}^c$ to $\text{PE}_{\text{ROI}_{r}^i}$ to indicate ROI level. In summary, we create region-specific position embeddings to differentiate between patches from distinct areas of the image. An example is shown in Fig.~\ref{fig:rspe}.

\subsection{Masked Recognition Model (MRM)}
\label{sec:mrb}
For each image, Retina Patch results in different numbers of tokens because different ROIs create different areas of intersection. For example, the number of patches from $\text{ROI}_0$ in Fig.~\ref{fig:rspe} is $12\times 12$ but the upper torso $\text{ROI}_1$ subtracts $4 \times 4$ patches from $\text{ROI}_0$ to avoid overlap. This operation leads to a different number of tokens per image, which prevents us from training and testing with batched inputs. 
To address the  token inconsistency, we propose the Masked Recognition Model (MRM), introducing two key techniques: (1) masking with attention scaling and (2) a variable masking rate. 

\begin{figure}
    \centering
    \includegraphics[width=1.0\linewidth]{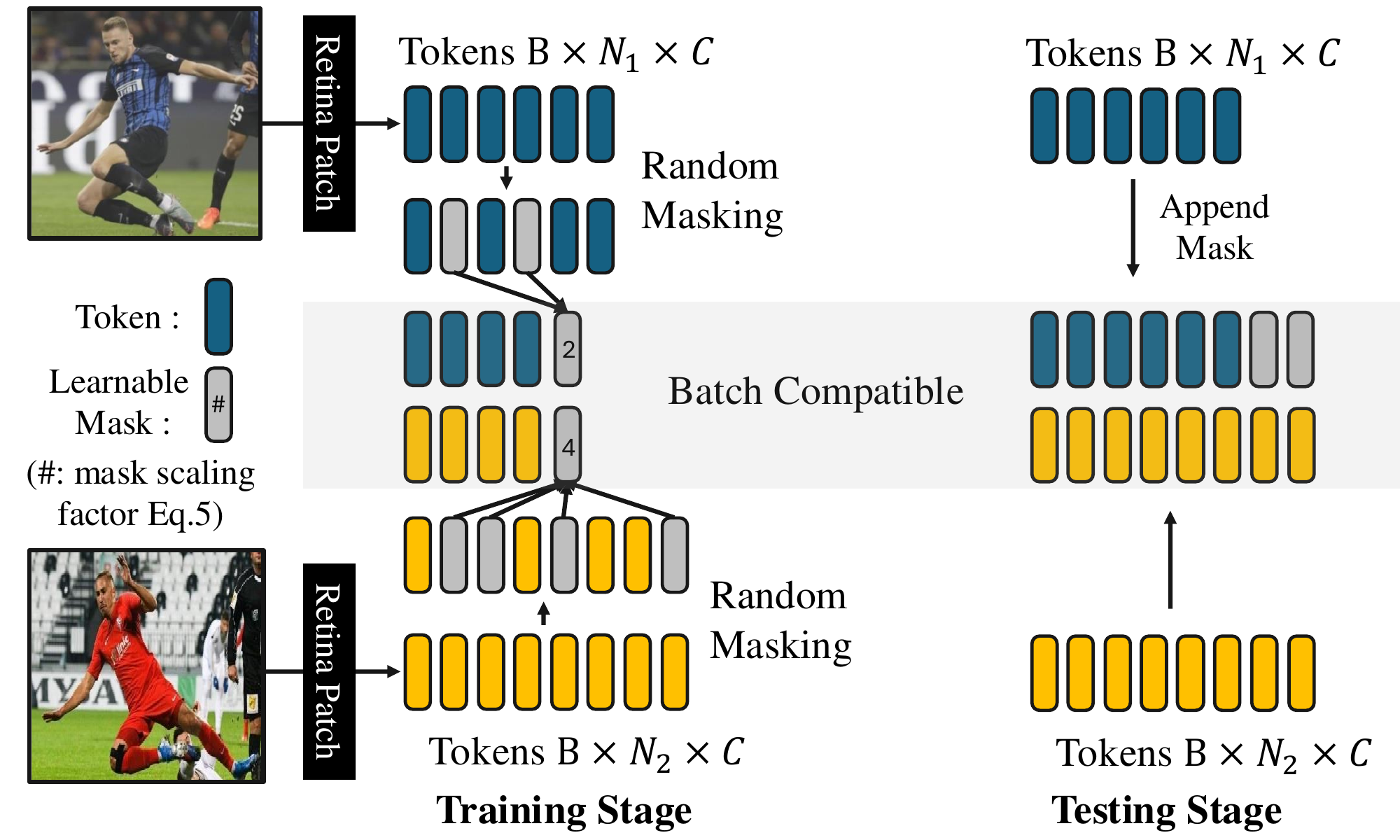}
    \vspace{-4mm}
    \caption{Illustration of Masked Recognition Backbone with masking and attention scaling trick for batched input during training. In testing, we pad with mask tokens to make the length the same. \vspace{-3mm}}
    \label{fig:mrm}
    
\end{figure}

\Paragraph{Masking with Attention Scaling} During training, we select tokens to keep. Unlike MAE~\cite{he2022masked}, which discards the masked tokens, we replace them with a learnable mask token. We do this because (i) the mask token will be used during testing for padding the input, and (ii) this allows the model to explicitly know \textit{how many} tokens are masked. Yet, since all masked tokens share the same value, we can reduce computation by applying the Attention Scaling Trick.

Specifically, although there are multiple masked tokens, we can achieve the same effect with a single mask token by adjusting its attention scores to reflect the total number of masked tokens. Let \( n_i \) be the total number of tokens for $i$-th image, \( n_k \) be the number of tokens we keep, and \( n_{m,i} = n_i - n_k \) be the number of masked tokens. We modify the attention computation in the Transformer as:
\begin{equation}
A = \text{softmax} \left( \mathbf{Q} \mathbf{K}^\top / \sqrt{d} + \boldsymbol{\delta} \right),
\end{equation}
where \( \mathbf{Q} \in \mathbb{R}^{(n_k+1) \times d} \) and \( \mathbf{K} \in \mathbb{R}^{(n_k+1)  \times d} \) are the query and key matrices with tokens to keep and one mask token. \( d \) is the embedding dimension. We add a bias matrix  \( \boldsymbol{\delta} \in \mathbb{R}^{n \times n} \) so that it is mathematically equivalent to repeating the mask tokens $n_{m,i}$ times during attention computation.
\begin{equation}
\boldsymbol{\delta}_{ij} =
\begin{cases}
\log n_{m,i}, & \text{if } j \text{ is the mask token}, \\
0, & \text{otherwise}.
\end{cases}
\end{equation}

In summary, we reduce the number of tokens from $n_i$ to $(n_k+1)$. Note that $(n_k+1)$ is fixed and not image dependent. But we adjust the attention to make it equivalent to using $n_i$ tokens where $n_{m,i}$ tokens are replaced by learnable mask tokens (proof in the Supp.). 
By applying the Attention Scaling Trick, we handle varying token counts in training. Also in practice, $n_k$ is set to be about $1/3$ of $n_i$, masking 66\% of tokens for the speed gain.  During testing, we simply find the longest token length and pad the others with the mask token to batchify the inputs. An illustration is in Fig.~\ref{fig:mrm}. 

\Paragraph{Variable Masking Rate} 
As we view masked training as a form of augmentation, we randomize $n_k$ during training and adjust the batch size correspondingly. 
For each batch, let \(\hat{n}_k\) be the sampled number of tokens to keep,
\begin{equation}
\hat{n}_k = n_k + (n_i - n_k) \cdot e^{-\lambda \cdot U(0, 1)}.
\label{eq:sample}
\end{equation}
\(\lambda\) is a scaling factor, and \(U(0, 1)\) denotes a random uniform distribution between 0 and 1. In short, \(\hat{n}_k\) is sampled from a distribution that peaks at \(n_k\) and exhibits an exponential decay in probability toward \(n_i\) (see Supp. for its visualization).

With a randomized token length \(n_k\), we adjust the batch size \(B\) based on the relationship \( n_k^2 \propto \frac{1}{B} \), where increasing \(n_k\) would require decreasing \(B\) to maintain the same GPU memory and FLOP. 
And we adjust the learning rate according to the effective batch size $\mathcal{L}_{\text{adj}} = \mathcal{L}_{\hat{n}_k} \times B_{\hat{n}_k} / B_{{n_k}}$ 
to maintain consistent gradient magnitudes per sample. 

The effect of (1) masking with attention scaling and (2) variable masking rate is ablated in Tab~\ref{tab:ablation}. While (1) and (2) are both helpful, the effect of (2) is more pronounced. 

\subsection{Semantic Attention Head (SAH)}

In biometric recognition, the head module is key for converting the backbone’s output feature map into a compact feature vector for recognition. Face recognition models flatten the feature map and apply linear layers~\cite{deng2019arcface,kim2022adaface}, while body recognition models use horizontal pooling~\cite{openpose,ye2024biggait}. However, these approaches rely on input image alignment (aligned face or standing body) which fails when there are large pose variations. 
To tackle this, we introduce a Semantic Attention Head (SAH) that extracts semantic part features from key body parts, making the representation less sensitive to pose.

\begin{figure}[t]
    \centering
    \includegraphics[width=1.0\linewidth]{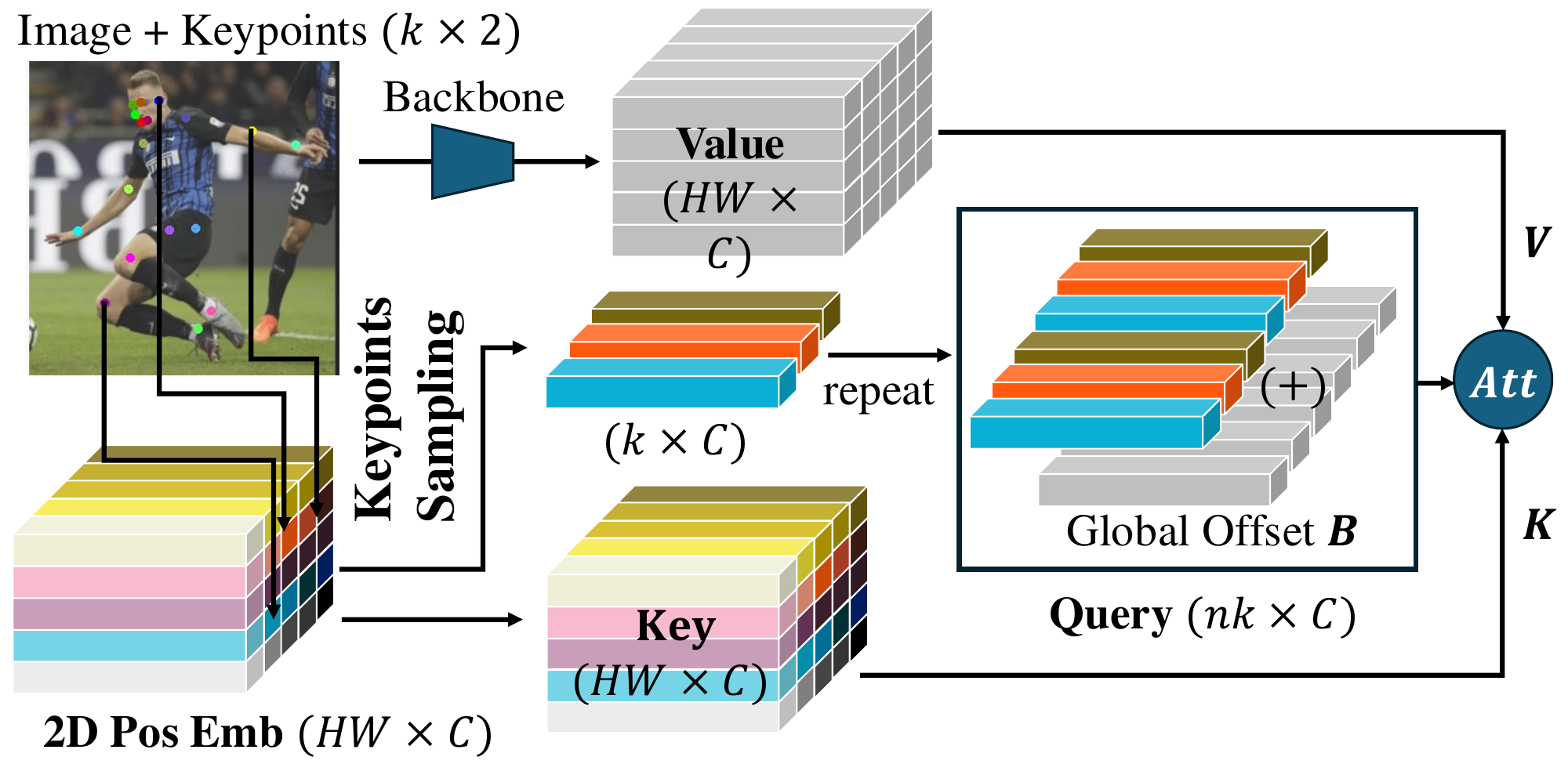}
    \vspace{-5mm}
    \caption{
Illustration of semantic pooling in Semantic Attention Head. Keypoints (e.g., nose, feet) are used to grid-sample position embeddings (PE), forming queries that repeat 
$n$ times and added with a global offset bias $\mathbb{B}$. This setup enables attention to slightly varied locations around each keypoint. Value comes from ViT backbone and Key is the PE. Result is a learned pooling mechanism. \vspace{-4mm}}
\label{fig:semantic_head}
\end{figure}

Our method uses keypoints ({\it e.g.}, nose, hip) for capturing semantic parts. But instead of sampling features only at keypoints, which may miss the surrounding context, SAH \textit{learns} to pool features around each keypoint. We construct a semantic query \( \mathbf{Q}_{kp}^i \) ({\it e.g.}, nose) using 2D position embeddings (PE) from the backbone, sampled at keypoint locations:
\begin{equation}
    \mathbf{Q}_{kp}^i = \text{GridSample}(\text{PE}, \text{kp}^i) + \mathbf{B},
    \label{eq:sah}
\end{equation}
where $\text{PE}$ is the fixed 2D image position embedding. $\text{kp}_i \in \mathbb{R}^{nk\times 2}$ is the image-specific predicted keypoints~\cite{openpose}. We duplicate keypoints $n$ times and add shared bias $\mathbf{B} \in \mathbb{R}^{nk \times C}$. The purpose of $\mathbf{B}$ is to learn to offset the center of attention so that it learns to pool from diverse locations around keypoints. 
 Key in attention is the fixed PE. Value is the backbone’s feature map. The attenton~\cite{ye2024differential} with $\mathbf{Q}_{kp}^i$ captures the neighborhood of the backbone feature map around keypoints:
    \begin{equation}
    \mathbf{O}_{\text{part}}^i = \text{Attention}\left(\mathbf{Q}_{kp}^i, \text{PE}, \text{backbone}(\mathbf{X}^i)\right).
\label{eq:sah_attention}
\end{equation}
The \( \mathbf{O}_{\text{part}}^i \in \mathbb{R}^{B \times k \times C} \) contains semantic part features corresponding to \( k \) keypoints. Finally, applying a multi-layer perceptron (MLP) to the flattened \( \mathbf{O}_{\text{part}}^i \) produces a  feature,
\begin{equation}
    f^i = \text{MLP}(\text{flatten}(\mathbf{O}_{\text{part}}^i)).
\end{equation}
By learning to pool features adaptively around each keypoint, this attention mechanism enables pose-invariant recognition that goes beyond conventional alignment-dependent methods. Fig.~\ref{fig:semantic_head}  illustrates the attention pooling.

\begin{table*}[t!]
\centering
\scriptsize
\renewcommand{\arraystretch}{0.94} 
\begin{tabular}{c|c|l|c|cccccccccc}
\hline
\multirow{2}{*}{Method} & \multirow{2}{*}{Arch} & \multirow{2}{*}{Train Data} & \multirow{2}{*}{Avg} & \multicolumn{2}{c}{LTCC (General)} & \multicolumn{2}{c}{PRCC (SC)~\cite{yang2019person}} & \multicolumn{2}{c}{CCVID (General)} & \multicolumn{2}{c}{Market1501} & \multicolumn{2}{c}{MSMT17~\cite{msmt17}} \\
 & &  &  & top1 & mAP & top1 & mAP & top1 & mAP & top1 & mAP & top1 & mAP \\
\hline
CAL~\cite{gu2022clothes} & R50 & LTCC & 48.64 & 74.04 & 40.84 & 99.51 & 95.64 & 75.63 & 28.08 & 35.60 & 16.11 & 15.92 & 5.06 \\
CAL~\cite{gu2022clothes} & R50 & PRCC & 35.07 & 20.69 & 6.19 & 100.00 & 99.76 & 74.48 & 20.86 & 18.97 & 6.47 & 2.56 & 0.69 \\
CAL~\cite{gu2022clothes} & R50 & LTCC+PRCC & 49.69 & 72.41 & 38.12 & 99.54 & 99.01 & 74.83 & 29.43 & 43.65 & 21.03 & 14.48 & 4.44 \\
CLIP3DReID~\cite{liu2024distilling} & R50 & LTCC & 50.89 & \textbf{75.66} & \textbf{45.15} & 99.43 & 96.43 & 77.28 & 30.01 & 41.66 & 20.33 & 17.45 & 5.50   \\
CLIP3DReID~\cite{liu2024distilling} & R50 & PRCC & 35.14 & 21.30 & 6.19 & 100.00 & \textbf{99.84} & 71.73 & 19.81 & 20.93 & 7.49 & 3.28 & 0.85 
\\\hdashline
SOLDIER~\cite{chen2023beyond} & Swin-Base & LU4M+Market1501 & 64.85 & 73.83 & 36.28 & 99.51 & 99.53 & 40.27 & 36.56 & \textbf{97.03 }& \textbf{94.04} & 48.64 & 22.77 \\
SOLDIER~\cite{chen2023beyond} & Swin-Base & LU4M+MSMT17 & 70.19 & 74.44 & 36.74 & 99.30 & 98.71 & 32.73 & 27.76 & 89.85 & 73.20 & \textbf{91.12} & \textbf{78.01} \\\hdashline
HAP~\cite{yuan2023hap} & ViT-Base & LU4M+LTCC & 45.71 & 65.11 & 29.02 & 95.53 & 86.44 & 44.16 & 30.43 & 51.63 & 27.29 & 20.89 & 6.56 \\
HAP~\cite{yuan2023hap} & ViT-Base & LU4M+PRCC & 54.09 & 63.29 & 29.36 & 98.84 & 98.38 & 49.15 & 37.73 & 73.49 & 50.11 & 29.61 & 10.99 \\
HAP~\cite{yuan2023hap} & ViT-Base & LU4M+Market1501 & 66.61 & 73.02 & 35.97 & 99.30 & 98.45 & 54.74 & 45.14 & 96.23 & 92.20 & 48.01 & 23.02 \\
HAP~\cite{yuan2023hap} & ViT-Base & LU4M+MSMT17 & 66.64 & 67.95 & 32.07 & 99.15 & 96.50 & 37.81 & 30.52 & 80.37 & 57.07 & 89.13 & 75.85 \\
HAP~\cite{yuan2023hap} & ViT-Base & WebBody4M (Ours) & 61.49 & 56.80 & 25.88 & 99.72 & 98.26 & 89.00 & 71.65 & 66.18 & 42.41 & 43.61 & 21.42 \\\hdashline
SapiensID (Ours) & ViT-Base & WebBody4M (Ours) & \textbf{73.05} & 72.01 & 34.56 & \textbf{100.00} & 98.79 & \textbf{92.57 }&\textbf{ 77.82} & 88.18 & 68.26 & 67.25 & 31.02 \\
\hline
\end{tabular}
\vspace{1mm}
\textbf{(a) Short-Term ReID}

\begin{tabular}{c|c|l|c|cccccccccc}
\hline
\multirow{2}{*}{Method} & \multirow{2}{*}{Arch} & \multirow{2}{*}{Train Data} & \multirow{2}{*}{Avg} & \multicolumn{2}{c}{LTCC (CC)~\cite{shu2021large}} & \multicolumn{2}{c}{PRCC (CC)} & \multicolumn{2}{c}{CCVID (CC)~\cite{gu2022clothes}} & \multicolumn{2}{c}{CCDA~\cite{3DInvarReID}} & \multicolumn{2}{c}{Celeb-ReID~\cite{huang2019celebrities}} \\
 & &  &  & top1 & mAP & top1 & mAP & top1 & mAP & top1 & mAP & top1 & mAP \\
\hline
CAL~\cite{gu2022clothes} & R50 & LTCC & 28.40 & 38.01 & 18.84 & 37.00 & 35.20 & 74.97 & 25.08 & 3.91 & 9.67 & 37.42 & 3.92 \\
CAL~\cite{gu2022clothes} & R50 & PRCC & 24.71 & 6.38 & 3.14 & 55.69 & 55.64 & 71.61 & 17.40 & 2.85 & 8.61 & 23.59 & 2.20 \\
CAL~\cite{gu2022clothes} & R50 & LTCC+PRCC & 29.46 & 33.16 & 16.27 & 45.39 & 45.42 & 73.89 & 26.65 & 3.74 & 9.14 & 37.11 & 3.81 \\ 
CLIP3DReID~\cite{liu2024distilling} & R50 & LTCC & 30.24 & 41.84 & \textbf{22.58} & 40.81 & 38.38 & 76.28 & 26.69 & 4.31 & 10.18 & 37.31 & 4.02 \\
CLIP3DReID~\cite{liu2024distilling} & R50 & PRCC & 25.79 & 6.63 & 3.17 & 62.40 & 61.97 & 69.32 & 16.38 & 3.17 & 8.89 & 23.82 & 2.17
\\\hdashline
SOLDIER~\cite{chen2023beyond} & Swin-Base & LU4M+Market1501 & 24.84 & 25.00 & 12.18 & 26.87 & 32.12 & 39.61 & 35.48 & 8.62 & 16.48 & 46.37 & 5.66 \\
SOLDIER~\cite{chen2023beyond} & Swin-Base & LU4M+MSMT17 & 22.17 & 26.02 & 11.33 & 22.27 & 25.36 & 31.85 & 26.48 & 8.79 & 15.54 & 47.95 & 6.14 \\\hdashline
HAP~\cite{yuan2023hap} & ViT-Base & LU4M+LTCC & 20.21 & 25.00 & 11.63 & 26.14 & 22.34 & 41.64 & 25.77 & 4.56 & 11.18 & 30.28 & 3.54 \\
HAP~\cite{yuan2023hap} & ViT-Base & LU4M+PRCC & 26.12 & 29.08 & 12.52 & 38.05 & 41.94 & 45.73 & 33.12 & 5.13 & 13.40 & 37.79 & 4.48 \\
HAP~\cite{yuan2023hap} & ViT-Base & LU4M+Market1501 & 27.49 & 24.74 & 11.71 & 33.90 & 37.00 & 52.37 & 41.33 & 8.30 & 16.02 & 44.38 & 5.20 \\
HAP~\cite{yuan2023hap} & ViT-Base & LU4M+MSMT17 & 21.61 & 23.47 & 10.74 & 23.82 & 25.00 & 34.54 & 26.81 & 6.27 & 13.33 & 46.37 & 5.77 \\
HAP~\cite{yuan2023hap} & ViT-Base & WebBody4M (Ours)& 44.90 & 22.70 & 9.96 & 54.93 & 49.38 & 88.34 & 68.66 & 28.80 & 41.49 & 65.78 & 18.93 \\\hdashline
SapiensID (Ours) & ViT-Base & WebBody4M (Ours) & \textbf{66.30} & \textbf{42.35} & 17.79 & \textbf{78.75} & \textbf{72.60} & \textbf{88.72} & \textbf{72.22} & \textbf{61.84} & \textbf{69.08} & \textbf{92.80} & \textbf{66.92 }\\
\hline
\end{tabular}
\vspace{1mm}
\textbf{(b) Long-Term ReID}
\vspace{-2mm}
\caption{Generalization comparison with SoTA ReID models on two settings. 
"Long-term" refers to clothing change (CC) protocol of LTCC, PRCC, and CCVID datasets, while "short-term" the same clothing (SC) protocol. 
For other datasets, the data capture characteristics define short or long-term conditions. 
SapiensID demonstrates superior generalization in both settings. 
Our WebBody4M dataset shows higher performance in long-term ReID, but not with the dataset alone,
as shown in the comparison of HAP vs SapiensID with the same training set. 
The proposed Retina-Patch and Semantic Attention Head are essential for learning under large pose and scale variations.}
\label{tab:wholebody}
\vspace{-3mm}
\end{table*}

\Paragraph{Training with Mixed Datasets} 
While SAH effectively handles pose variations, we hypothesize that key cues for recognition differ between short-term and long-term training datasets. Clothing and hairstyle, for example, are useful in short-term datasets but less reliable in long-term due to possible appearance changes. 

To aid learning with mixed datasets which combines short-term and long-term datasets, we introduce one more measure during training. We introduce a learnable scale that controls the importance of individual part features in $(\mathbf{O}_{\text{part}}^i)$ for each dataset. It is to allow the model to emphasize features that are most discriminative for each dataset. During testing, however, we can use the average scale because we do not want to utilize the knowledge about the test dataset a priori. 

Specifically, let $\mathbf{W}_t \in \mathbb{R}^{k}$ be a weight for the $t$-th dataset. For each sample, we choose the weight and apply
\begin{equation}
f^i = \text{MLP}(\text{flatten}(\mathbf{O}_{\text{part}}^i \cdot \sigma(\mathbf{W}_t)) ),
\label{eq:weights}
\end{equation}
 where $\sigma$ is the Sigmoid function, ensuring weights are between 0 and 1, controlling the influence of each of the $k$ semantic parts. 
 We observe that after training, short-term datasets tend to focus on the clothing and long-term datasets focus on the upper torso. 
 The learned weight is visualized in Supp. 
 The weight is for learning discriminative parts during training but we do not use dataset-specific weights in testing.

\subsection{WebBody Dataset}
\label{sec:webbody}
To facilitate the training, we collect a large-scale, labeled human dataset from the web. 
Specifically, we gather $94$ million images with $3.8$ million celebrity names. 
Given the inherent noise in web-sourced name queries, we perform extensive label cleaning. 
First, we use YOLOv8~\cite{yolov8_ultralytics} to crop the dominant person in each image to a size of $384 \times 384$, adding padding to maintain aspect ratio. 
We then extract facial features using RetinaFace~\cite{deng2020retinaface} and KP-RPE~\cite{kprpe}. 
Following the approach in~\cite{zhu2021webface260m}, we apply DBSCAN~\cite{dbscan} clustering to identify the most consistent group of images for each name. 
By assuming all images stem from a single name query, we relax the similarity threshold beyond conventional face recognition standards. 
We also exclude any images with face features matching those in validation sets~\cite{agedb,cfpfp,lfw,calfw,cplfw}.

This process yields a labeled dataset of $4.4$ million images from $217,722$ unique subjects. 
However, as the dataset is labeled based on facial similarity, it lacks images where the face is obscured ({\it e.g.}~back-facing images). 
Thus, we incorporate additional body ReID training datasets~\cite{shu2021large,yang2019person,huang2019celebrities,zheng2015scalable,sun2018beyond,luperson,gu2022clothes,xu2021deepchange}, which account for $\sim$10\% of the final dataset. 
The resulting dataset—named WebBody4M—comprises $4.9$ million images and $263,920$ subjects in total. 
WebBody4M is the largest labeled dataset to date with high pose and scale variation. The keypoint visibility distribution of different body parts in Supp. shows a predominance of visible upper body, with visibility decreasing gradually down the body (around $17\%$ visible ankles). 
An example of the WebBody4M dataset can be seen in Fig.~\ref{fig:one}.

The dataset collection and label cleaning procedure is similar to WebFace4M dataset~\cite{zhu2021webface260m}. 
We compare the face-cropped version of WebBody4M with WebFace4M and observe that an FR model trained on WebBody4M-FaceCrop is similar in performance to WebFace4M
(see details in the Supp.).  Separate from the WebBody4M, we also prepare a test set called WebBody-test to evaluate the cross pose-scale ReID performance. 
It comprises $96,624$ images of $4,000$ gallery and probe subjects. Examples are shown in Fig.~\ref{fig:fig2}.


\begin{table}[t!]
\centering
\scriptsize
\renewcommand{\arraystretch}{0.94}
\scalebox{0.95}{
\begin{tabular}{c|c|c|c|cc}
\hline

\multirow{2}{*}{Method} & \multirow{2}{*}{Arch} & \multirow{2}{*}{Train Data} & \multirow{2}{*}{Avg} & \multicolumn{2}{c}{WebBody Testset} \\
 & &  &  & top1 & mAP \\
\hline
CAL~\cite{gu2022clothes} & R50 & PRCC & 2.47 & 4.29 & 0.64 \\
CAL~\cite{gu2022clothes} & R50 & LTCC & 3.79 & 6.57 & 1.02 \\\hdashline
SOLDIER~\cite{chen2023beyond} & Swin-Base & Market1501 & 3.22 & 5.42 & 1.02 \\
SOLDIER~\cite{chen2023beyond} & Swin-Base & MSMT17 & 5.96 & 9.95 & 1.98 \\\hdashline
HAP~\cite{yuan2023hap} & ViT-Base & LTCC & 1.74 & 2.89 & 0.58 \\
HAP~\cite{yuan2023hap} & ViT-Base & PRCC & 2.61 & 4.37 & 0.85 \\
HAP~\cite{yuan2023hap} & ViT-Base & Market1501 & 4.31 & 7.22 & 1.39 \\
HAP~\cite{yuan2023hap} & ViT-Base & MSMT17 & 4.87 & 8.22 & 1.52 \\
HAP~\cite{yuan2023hap} & ViT-Base & WebBody4M & 47.12 & 64.36 & 29.89 \\\hdashline

SapiensID (Ours) & ViT-Base & WebBody4M & \textbf{64.41} & \textbf{76.82} & \textbf{52.00} \\
\hline

\end{tabular}
}
\caption{ReID Performance on variable pose and scale settings. }
\vspace{-3mm}
\label{tab:webbody}
\end{table}

\section{Experiments}
\Paragraph{Implementation Details}
To train SapiensID on Webbody4M, we use AdaFace~\cite{kim2022adaface} loss and ViT-Base with KP-RPE as the main backbone~\cite{kprpe}, following the convention of face recognition model training pipeline. 
We do not include additional losses such as Triplet Loss~\cite{schroff2015facenet} since there are a sufficient number of subjects in the training set.
Input image size is $384\times 384$ with white padding if the aspect ratio is not $1$. 
We use $3$ ROIs (whole image, upper torso, and head) and the grid size per ROI is $12\!\times\!12$ leading to a maximum $144\!\times\!3$ number of patches. 
With masked recognition training, we replace at most $66\%$ of tokens with mask (Sec.~\ref{sec:mrb}), leading to  $\sim$$9$ times speed up in training. 
The masking probability and batch size rule are discussed in Supp. 
We use $7$ H100 GPUs to train the whole model in 2 days, starting from scratch.


\Paragraph{Whole Body ReID} The task identifies individuals walking or standing in distant camera views, categorized into short or long-term scenarios based on the time gap between captures and the likelihood of clothing changes. Tab.~\ref{tab:wholebody} shows our results on the ReID benchmarks. A significant departure from prior works is the use of a single SapiensID model across all evaluation settings, whereas previous methods employ fine-tuned models for each evaluation dataset (one model per dataset). 
This distinction highlights SapiensID’s potential for deployment in diverse, unseen, real-world environments.

SapiensID achieves the highest average mAP of 73.05\% across short-term ReID benchmarks. 
Furthermore, we attain SoTA results on all evaluated long-term ReID datasets. 
This strong performance underscores the value of the WebBody4M dataset in training a generalizable model. 
However, this achievement would not have been possible without our SapiensID architecture, which effectively handles variations in pose and visible body areas. 
A strong baseline (HAP~\cite{yuan2023hap}) trained on WebBody4M alone does not achieve comparable results, highlighting the importance of our architectural innovations to leverage the dataset. 
SapiensID marks a significant advance by being the first single model capable of strong performance across short and long-term ReID tasks.

\begin{table}[t]
\centering
\scriptsize
\renewcommand{\arraystretch}{0.94}
\begin{tabular}{c|c|cc}
\hline
 \multirow{2}{*}{Method}  &  \multirow{2}{*}{Training Data  }  & \multicolumn{2}{c}{OccludedReID} \\ 
       &             & top1 & mAP \\ \hline
KPR~\cite{somers2025keypoint} + SOLDIER & LU4M +OccludedReID   & 84.80 & \textbf{82.60} \\
SapiensID & WebBody4M & \textbf{87.30} & 75.57 \\ \hline
\end{tabular}
\caption{Performance in occluded ReID. SapiensID achieves a higher top-1 accuracy, while KPR~\cite{somers2025keypoint} shows a higher mAP. SapiensID is trained without OccludedReID training data. }
\label{tab:occlusion}
\end{table}

\begin{table}[t!]
    \centering
    \scriptsize
    \renewcommand{\arraystretch}{0.94}
    \begin{tabular}{c|cc}
    \hline
        Method & AdaFace-ViT~\cite{kim2022adaface} & SapiensID (Ours) \\ \hline
        Train Data & \parbox[c]{2.5cm}{\centering WebBody4M-FaceCrop} & WebBody4M \\ \hline
        LFW~\cite{lfw} & 99.82 & 99.82 \\ 
        CPLFW~\cite{cplfw} & 95.12 & 94.85 \\ 
        CFPFP~\cite{cfpfp} & 99.19 & 98.74 \\ 
        CALFW~\cite{calfw} & 96.07 & 95.78 \\ 
        AGEDB~\cite{agedb} & 97.97 & 97.33 \\ \hdashline
        Face Avg & \textbf{97.63} & 97.31 \\ \hline
        LTCC~\cite{shu2021large}  & 21.70 & 72.01 \\ 
        Market1501~\cite{market}  & 7.81 & 88.18 \\ \hdashline
        Body Avg & 14.76 & \textbf{80.10}
        \\ \hline
        Combined Avg & 56.19 & \textbf{89.80} \\ \hline
    \end{tabular}
    \caption{Performance on cross-modality setting. Face recognition is evaluated on aligned face recognition datasets and body recognition is evaluated on short-term ReID datasets. LTCC and Market1501 measure top1 of short-term setting. \vspace{-3mm}}
    \label{tab:crossmodality}
\end{table}

\Paragraph{Cross Pose-Scale ReID} Real-world human recognition can present scenarios where subjects are captured across varying camera viewpoints and exhibit diverse poses, such as sitting, bending, or engaging in activities. For example, a security camera might capture a person standing upright, while a social media photo shows the same individual sitting in a cafe. This poses a challenge for conventional ReID systems. We refer to this setting as Cross Pose-Scale ReID.

To evaluate this setting, we introduce the WebBody-Test dataset, specifically designed to encompass such pose and scale variations. Tab.~\ref{tab:webbody} details the performance comparison on this dataset. 
Conventional ReID models struggle to generalize to this scenario due to the significant shift in visual appearance caused by pose and scale changes. SapiensID with the highest performance establishes a strong baseline for this research area. Since the task itself is challenging, there is still room for improvement. WebBody dataset demonstrates the potential of SapiensID to address the complexities of Cross Pose-Scale ReID, while offering a valuable starting point for future research in this area.

\Paragraph{Occluded ReID} 
Occlusions, whether due to obstacles in the scene or self-occlusion from the subject's pose, present a further challenge for robust human recognition. 
We evaluate SapiensID in occluded scenarios on the OccludedReID dataset~\cite{zhuo2018occluded}, comparing with KPR~\cite{somers2025keypoint}, a SoTA method designed for occlusion handling. 
As shown in Tab.~\ref{tab:occlusion}, SapiensID achieves a competitive performance of top-1 87.30\%, demonstrating its strong ability to handle occlusions even without being explicitly trained on the OccludedReID dataset. 
This result further underscores the value of our architecture and training dataset in learning representations that are resilient to real-world challenges like occlusions.

\begin{table}[!t]
    \centering
    \scriptsize
    \renewcommand{\arraystretch}{0.94}
    \begin{tabular}{l|c|ccc}
    
    \hline
         & \multirow{2}{*}{All} & \multirow{2}{*}{Face} & \multicolumn{2}{c}{Whole Body ReID} \\
        &  &  & Short & Long \\ \hline
        (1) ViT & 59.54 & 90.63 & 56.17 & 31.81 \\ 
        (2) ViT+RP & 66.35 & 92.93 & 59.16 & 46.95 \\ 
        (3) ViT+SAH & 71.67 & 95.84 & 72.63 & 46.55 \\ 
        \multicolumn{1}{l|}{\makecell[l]{(4) ViT+RP+SAH (SapiensID)}} & \textbf{78.67} & \textbf{96.66} & \textbf{73.05} & \textbf{66.30} \\ \hline
        (4) $-$ Learned Mask & 76.99 & 96.08 & 70.44 & 64.46 \\ 
        (4) $-$ Variable $n_k$ & 74.39 & 95.95 & 69.58 & 57.64 \\ \hline
    \end{tabular}
\caption{Ablation study of SapiensID. 
Face is the average accuracy of CPLFW, CFPFP, CALFW, and AGEDB. 
Short and Long Term use the average of the datasets in Tab~\ref{tab:wholebody}. 
Results show the necessity and strong complementarity of both RP and SAH in SapiensID.}
\label{tab:ablation}
\end{table}

\begin{table}[t!]
\centering
\scriptsize
\renewcommand{\arraystretch}{0.94}
\begin{tabular}{r|l|cccc}
\hline
\textbf{} & \textbf{}  & \multicolumn{2}{c}{LTCC CC}   & \multicolumn{2}{c}{PRCC CC}  \\
\textbf{} & \textbf{}  & Top1  & mAP   & Top1  & mAP   \\\hline
1        & \textcolor{white}{0$-$}None    & 0.00  & 3.56  & 1.47  & 4.28 \\
2        & 1$+$Nose      & 25.77 & 5.78  & 27.21 & 21.04 \\
3         & 2$+$Eye    & 30.61 & 8.87  & 63.87 & 55.17 \\
4         & 3$+$Mouth      & 38.01 & 11.81 & 73.36 & 65.05 \\
5         & 4$+$Ear & 39.80 & 14.05 & 77.65 & 70.45 \\
6         & 5$+$Shoulder    & 41.84 & 15.82 & 79.73 & 73.14 \\
7         & 6$+$Elbow    & 41.07 & 16.64 & \textbf{80.55} & \textbf{73.54} \\
8         & 7$+$Wrist      & 41.07 & 17.16 & 79.34 & 73.16 \\
9         & 8$+$Hip     & 40.56 & 17.50 & 79.99 & 73.38 \\
10         & 9$+$Knee    & 42.35 & 17.73 & 79.00 & 72.88 \\
11         & 10$+$Ankle (Full)       & \textbf{42.35} & \textbf{17.79} & 78.75 & 72.6  \\\hline
\end{tabular}
\caption{Impact of adding body parts on ReID. None means all features are zeroed out. Each row adds features to the previous row. \vspace{-6mm} }
\label{tab:part}
\end{table}

\Paragraph{Face Recognition} We evaluate on traditional aligned face recognition benchmarks to assess the ability to handle FR tasks. Tab.~\ref{tab:crossmodality} compares SapiensID with a SoTA FR model, AdaFace~\cite{kim2022adaface}, both with a ViT-Base backbone. 
AdaFace is trained on faces aligned and cropped to $112\times 112$ by~\cite{deng2020retinaface}. 
AdaFace achieves a slightly higher average accuracy of 97.63\% across five benchmarks. 
This marginal difference is expected, given AdaFace's training on tightly cropped, aligned faces. 
However, SapiensID's performance remains highly competitive,
bridging the gap between specialized face recognition and general human recognition tasks.

While AdaFace excels in FR datasets, its performance degrades when applied to ReID datasets which contain images without visible face region (\textit{e.g.} back of the head). 
AdaFace is evaluated by cropping faces using~\cite{deng2020retinaface}. 
In contrast, SapiensID maintains strong performance across both modalities. More experiments can be found in Supp B.10 and B.11.

\label{sec:ablation}
\Paragraph{Ablation of Components} Tab.~\ref{tab:ablation} ablates SapiensID's key components: Retina Patch (RP) and Semantic Attention Head (SAH). 
Starting from a simple ViT backbone with AvgMax pooling~\cite{gu2022clothes} as a baseline, we progressively incorporate RP and SAH to analyze their individual and combined contributions. 
Performance is evaluated across face recognition and both short-term and long-term ReID. 
The results show that both RP and SAH are essential.

We also show the importance of MRM. (4) - Learned Mask means using MAE~\cite{he2022masked} to simply drop tokens. (4) - Variable $n_k$ is fixing $n_k$ without sampling. The result shows that learned mask is of some benefit while changing the masking rate during training is of larger benefit.

\begin{figure}
    \centering
    \includegraphics[width=1.0\linewidth]{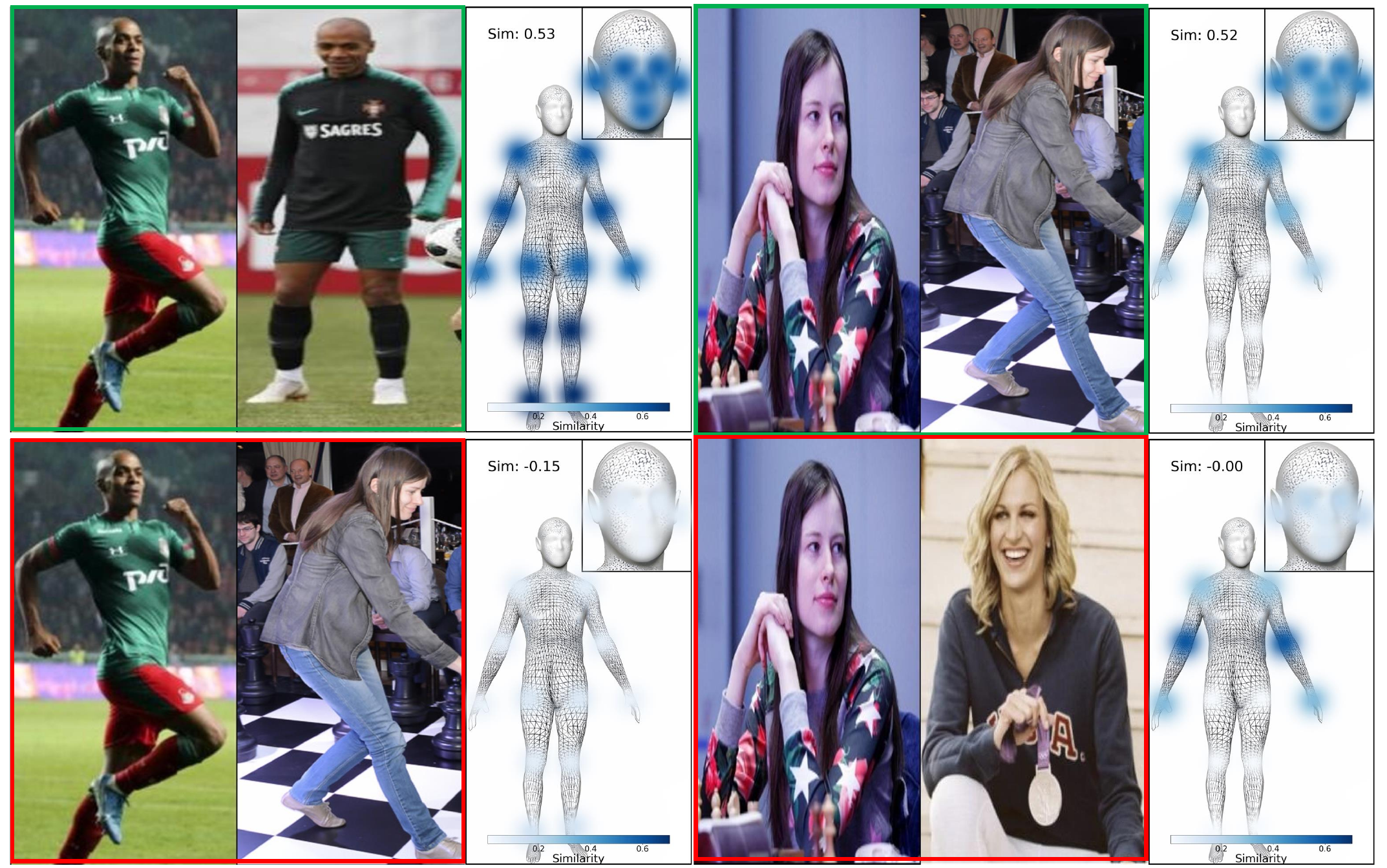}
    \vspace{-3mm}
    \caption{Part Similarity Visualization. Top shows the same subject pairs. Bottom shows different subject pairs. Part features provide some indication of where the similar parts are, but the final similarity is generated through a nonlinear mapping of the part features. \vspace{-3mm}}
    \label{fig:partsim}
    
\end{figure}

 \Paragraph{Analysis of Part Contribution} To see the impact of body parts in recognition, we erase part features by making them zero. Tab.~\ref{tab:part} shows a trend of performance gain as more parts are added. For LTCC dataset accuracy increases from 25.77\% to 42.35\% as body parts from the nose to ankle are incorporated. This suggests that including the full range of body parts aids recognition. In contrast, PRCC achieves high performance by using upper body cues, reaching a top-1 accuracy of 80.55\% with parts up to the shoulder and elbow. Lower body features add minimal or even negative value. This analysis implies the benefit of scenario-specific adjustments where relevant body regions can optimize recognition performance. We also visualize the part features similarity with sample images from the test set of WebBody4M in Fig~\ref{fig:partsim}. Samples of different scales and poses are visualized. 

\section{Conclusion}
SapiensID presents a paradigm shift in human recognition, moving beyond modality-specific models to a unified architecture capable of identification across diverse poses and body-part scales. Retina Patch, Semantic Attention Head, and Masked Recognition Model combined with WebBody4M dataset, enable SapiensID to achieve SoTA performance across various ReID benchmarks and establish a strong baseline for Cross Pose-Scale ReID. This work marks a step towards holistic human recognition systems. We include an in-depth discussion of the ethical impacts in Supp, ensuring that our approach respects intellectual property, privacy, and responsible data use.


\clearpage
\Paragraph{Acknowledgments} This research is based upon work supported by the Office of the Director of National Intelligence (ODNI), Intelligence Advanced Research Projects Activity (IARPA), via 2022-21102100004. The views and conclusions contained herein are those of
the authors and should not be interpreted as necessarily representing the official policies, either expressed or implied, of ODNI, IARPA, or the U.S. Government. The U.S. Government is authorized to reproduce and distribute reprints for governmental purposes notwithstanding any copyright annotation therein.
{
\small
\bibliographystyle{ieeenat_fullname}
\bibliography{bibliography}
}


\clearpage
\maketitlesupplementary

\appendix

\setcounter{page}{1}
\setcounter{section}{0}

\begin{figure*}[t!]
    \centering
    \includegraphics[width=0.8\linewidth]{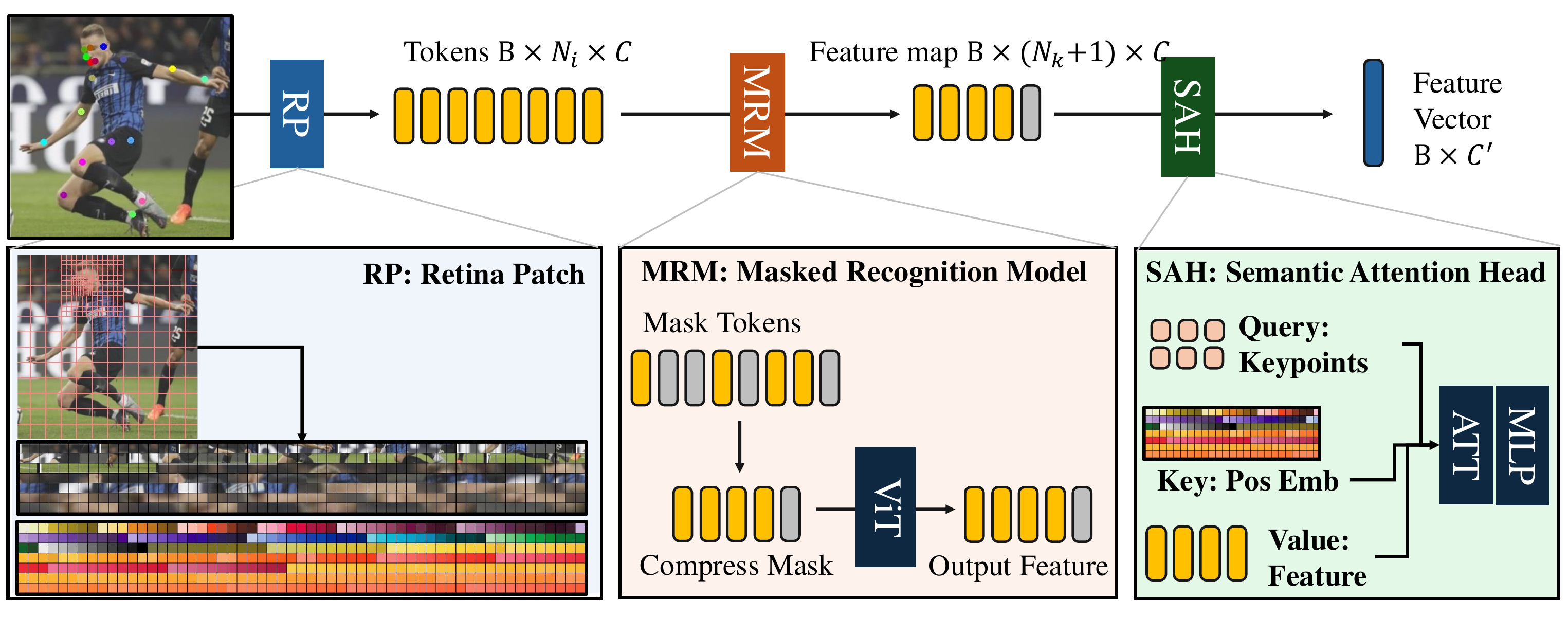}
    \caption{Illustration of the feature vector generation in SapiensID. First, Retina Patch (RP) generates image patches. Then, Masked Recognition Model (MRM) modifies the number of tokens. Finally, Semantic Attention Head (SAH) produces the feature vector from the set of tokens.}
    \label{fig:overview}
\end{figure*}

\section{Method Details}

\subsection{Training Details}
The training pipeline of SapiensID is largely similar to the setting of training a ViT model in face recognition~\cite{kprpe}. This is possible because WebBody4M is a labeled dataset with a sufficient number of subjects, just as face recognition datasets. We use the AdaFace~\cite{kim2022adaface} loss and optimize the model with the AdamW~\cite{adamw} optimizer for 33 epochs. The learning rate is scheduled by the Cosine Annealing Learning Rate Scheduler~\cite{loshchilov2016sgdr} with an additional warm-up period of 3 epochs. The maximum learning rate is set to $0.0001$. We use 7 A100 GPUs with a batch size of 128. We also change the classifier to PartialFC~\cite{an2021partial} with a sampling ratio of $0.1$ to save GPU memory and gain computation efficiency. Overview of the model is shown in Fig.~\ref{fig:overview}. 

For data augmentation, we find that it is important to use a moderate amount of geometric augmentation (zoom in-out: $0.9\sim1.1$, translation: $\pm0.05$) and aspect ratio adjustments ($0.95\sim1.05$). We also find it effective for improving aligned face recognition performance to include face-zoomed-in images frequently ($40\%$). We also oversample images that contain more visible keypoints because those images are relatively scarce (note Tab.~\ref{tab:keypoints}). 

\subsection{Notation Clarification in the Main Paper}

In Semantic Attention Pooling's SAH, the equation presented as Eq.~\ref{eq:sah_attention}:
\begin{equation}
    \mathbf{O}_{\text{part}}^i = \text{Attention}\left(\mathbf{Q}_{kp}^i, \text{PE}, \text{backbone}(\mathbf{X}^i)\right),
\end{equation}
$\text{Attention}(\mathbf{Q}, \mathbf{K}, \mathbf{V})$ is specifically defined as:
\begin{equation}
    \mathbf{O}_{\text{part}}^i = \text{softmax}\left(\frac{\mathbf{W}_q\mathbf{Q} \mathbf{W}_k\mathbf{K}^\top}{\sqrt{d}}\right) \mathbf{W}_v \mathbf{V},
\end{equation}
where \( \mathbf{Q} \), \( \mathbf{K} \), and \( \mathbf{V} \) represent the query, key, and value matrices, respectively, and \(\mathbf{W}_q \), \( \mathbf{W}_k \), and \( \mathbf{W}_v \) are their associated projection weights. This is how the size of the attenion is modulated during learning. 

Also notice that without the learnable projections  \( \mathbf{W}_{q,k,v} \) and a small \( d \), the attention simply focuses on the position with the highest proximity to the keypoint. To make sure that we have this feature from the sharp peak at the keypoint location, we additionally use
\begin{equation}
    \mathbf{O}_{\text{peak}}^i = \text{softmax}\left(\frac{\mathbf{Q} \mathbf{K}^\top}{\sqrt{d}}\right) \mathbf{V}.
\end{equation}

The final feature vector is computed by concatenating the two sets of semantic features $ \mathbf{O}_{\text{part}}^i$ and $ \mathbf{O}_{\text{peak}}^i$ and flattening them for MLP projection. Specifically, it is
\begin{equation}
f^i = \text{MLP}(\text{flatten}([\mathbf{O}_{\text{part}}^i, \mathbf{O}_{\text{peak}}^i]) ).
\label{eq:weights}
\end{equation}
The addition of $\mathbf{O}_{\text{peak}}^i$ is simply to ensure that the model always has the feature from the keypoint location. We have not tested how much performance gap is created by removing this inductive bias in SAH. The final number of part features is 152 (19 keypoints $\times$ 4 offset repeats $\times$ 2 from concatenating $\mathbf{O}_{\text{part}}^i$ and $ \mathbf{O}_{\text{peak}}^i$. 
We realize that the readers could be confused about the formulation of SAH attention, so we will make it clearer in the main paper.

\subsection{Things We Tried That Did Not Make it into the Main Algorithm}
\begin{itemize}
    \item We tried to initialize the model with the Sapiens~\cite{sapiens} pretrained backbone, thinking it would be a good starting point that leads to better generalization. However, it did not lead to better performance. We believe this is because: 1) our patch scheme is dramatically different from the original patch scheme, and 2) Sapiens is trained with the MAE~\cite{he2022masked} objective, which is suitable for dense prediction tasks. However, SapiensID is a classification (or metric learning) task. Dense prediction tasks prioritize spatial consistency and detailed reconstruction, whereas classification tasks focus on extracting discriminative features, which may require different feature representations.
    \item We tried using the differential layerwise learning rate~\cite{yang2019xlnet}, but it did not help and the learning was only slower. 
    \item We tried not learning the size and offset for the Semantic Attention Head (SAH) by simply taking the feature from the keypoint locations. This led to worse performance in general. 
\end{itemize}

\subsection{Transforming Keypoints to ROIs}

SapiensID relies on predicted keypoints to define Regions of Interest (ROIs). Assuming we have an input image roughly cropped around the visible body area (typically using a person detector's bounding box), we start with a set of predicted keypoints \(\mathbf{K} = \{(x_k, y_k)\}_{k=1}^N\), where \(N\) is the number of keypoints. Our goal is to generate bounding boxes for each ROI. Specifically, we generate two bounding boxes—for the face and the upper torso—in the format \((x_{1}, y_{1}, x_{2}, y_{2})\), representing the top-left and bottom-right corners.

\begin{enumerate}
    \item \textbf{Valid Keypoint Selection}:

    Let \(\mathcal{K} = \{1, 2, \dots, N\}\) be the set of keypoint indices. For each keypoint \(k \in \mathcal{K}\), the coordinates are \((x_k, y_k) \in \mathbb{R}^2\). We define a visibility indicator \(v_k\) for each keypoint:

    \begin{equation}
    v_k =
    \begin{cases}
    1, & \text{if } x_k \neq -1 \text{ and } y_k \neq -1, \\
    0, & \text{otherwise}.
    \end{cases}
    \end{equation}

    Define the sets of keypoint indices relevant to each ROI:
    
\quad

\begin{tabular}{ll}
\quad\quad\(K_1\): Left Eye            & \(K_6\): Left Mouth Corner \\
\quad\quad\(K_2\): Right Eye           & \(K_7\): Right Mouth Corner \\
\quad\quad\(K_3\): Left Ear            & \(K_8\): Left Shoulder \\
\quad\quad\(K_4\): Right Ear           & \(K_9\): Right Shoulder \\
\quad\quad\(K_5\): Nose                &  \\
\end{tabular}

\quad

Then Face Keypoints are 
\[
\mathcal{M}_{f} = \{ K_1, K_2, K_3, K_4, K_5, K_6, K_7 \}.
\]
And Upper Torso Keypoints are
\[
\mathcal{M}_{u} = \mathcal{M}_{f} \cup \{ K_8, K_9, K_{10}, K_{11} \}.
\]

    

    The valid keypoints for each ROI are those that are both visible and relevant:
    \begin{align}
    \mathcal{V}^{\text{face}} = \{ k \in \mathcal{M}_{f} \mid v_k = 1 \},\\
    \mathcal{V}^{\text{torso}} = \{ k \in \mathcal{M}_{u} \mid v_k = 1 \}.
    \end{align}

    \item \textbf{Bounding Box Center and Size Calculation}:

    For each ROI (face or upper torso), we compute the center using the set \(\mathcal{V}\), which is either \(\mathcal{V}^{\text{face}}\) or \(\mathcal{V}^{\text{torso}}\):

    First compute the minimum and maximum coordinates among valid keypoints:
        \begin{align}
        x_{\min} = \min_{k \in \mathcal{V}} x_k, \quad y_{\min} = \min_{k \in \mathcal{V}} y_k,\\
        x_{\max} = \max_{k \in \mathcal{V}} x_k, \quad y_{\max} = \max_{k \in \mathcal{V}} y_k.
        \end{align}

    Then calculate the center of the bounding box:
        \begin{equation}
        c_x = \frac{x_{\min} + x_{\max}}{2}, \quad c_y = \frac{y_{\min} + y_{\max}}{2}.
        \end{equation}
    Then determine the maximum distance \(d\) from the center to the valid keypoints:
        \begin{equation}
        d = \max_{k \in \mathcal{V}} \sqrt{(x_k - c_x)^2 + (y_k - c_y)^2}.
        \end{equation}

    \item \textbf{Bounding Box with Padding}:

First define the bounding box size \(s\) with a padding factor \(p\) (e.g., \(p = 0.3\)):
        \begin{equation}
        s = d \times (1 + p).
        \end{equation}

Then calculate the coordinates of the bounding box:
        \begin{align}
        x_{1} = c_x - s, \quad y_{1} = c_y - s,\\
        x_{2} = c_x + s, \quad y_{2} = c_y + s.
        \end{align}

    \item \textbf{Making Bounding Box Divisible}:
    To ensure that the patches cover the image without any overlap, the boundaries of the bounding box must \textit{snap} onto the patch grid. In other words, the bounding box coordinate should be divisible by the patch size ($p_w,p_h$) of the enclosing ROI. Let \(n_r\) and \(n_c\) be the desired number of rows and columns for patches within the ROI.  We modify the bounding box size \(s\) to ensure divisibility. 
\begin{align}
    x_1' &= \lfloor \frac{x_1}{p_w} \rfloor \times p_w, \quad 
    y_1' = \lfloor \frac{y_1}{p_h} \rfloor \times p_h \\
    x_2' &= \lceil \frac{x_2}{p_w} \rceil \times p_w,\quad
    y_2' = \lceil \frac{y_2}{p_h} \rceil \times p_h
\end{align}
The final, grid-aligned bounding box is then:
    \begin{equation}
    \mathbf{b} = (x_1', y_1', x_2', y_2') \in \mathbb{R}^4.
    \end{equation}
This snapping process ensures that the bounding box boundaries coincide with patch boundaries, resulting in clean, non-overlapping patch extraction. We compute two bounding boxes, $\mathbf{b}^{\text{face}}$ and $\mathbf{b}^{\text{torso}}$, using this process. All these steps can be conducted in GPU for efficient computation.
\end{enumerate}

\subsection{Proof of Scaled Attention Equivalence}

Let the scaled dot-product attention mechanism for self attention is defined as:
\[
A = \text{softmax}\left(\frac{\mathbf{Q}\mathbf{K}^\top}{\sqrt{d}}\right) \mathbf{V},
\]
We aim to prove that when a scaling factor $\boldsymbol{\delta} \in \mathbb{R}^{1 \times M}$ is added to the logits:
\[
A = \text{softmax}\left(\frac{\mathbf{Q}\mathbf{K}^\top}{\sqrt{d}} +  \boldsymbol{\delta}\right) \mathbf{V},
\]
this is equivalent to repeating each key $\mathbf{K}_j$ and value $\mathbf{V}_j$ exactly $m_j$ times, where $\delta_j = \log m_j$.

\noindent\textbf{Proof:} Consider the following term:
   \[
   \frac{\mathbf{Q}\mathbf{K}^\top}{\sqrt{d}} + \boldsymbol{\delta}.
   \]
   For a query $i$ and key $j$, the element of this matrix is:
   \[
   \left(\frac{\mathbf{Q}\mathbf{K}^\top}{\sqrt{d}} + \boldsymbol{\delta}\right)_{ij} = \frac{\mathbf{Q}_i \cdot \mathbf{K}_j^\top}{\sqrt{d}} + \log m_j,
   \]
   where $\mathbf{Q}_i$ is the $i$-th query and $\mathbf{K}_j$ is the $j$-th key.
   Applying the softmax function, we get:
   \[
   A_{ij} = \frac{\exp\left(\frac{\mathbf{Q}_i \cdot \mathbf{K}_j^\top}{\sqrt{d}} + \log m_j\right)}{\sum_k \exp\left(\frac{\mathbf{Q}_i \cdot \mathbf{K}_k^\top}{\sqrt{d}} + \log m_k\right)}.
   \]
   Using the property $\exp(a + b) = \exp(a)\exp(b)$, this simplifies to:
   \[
   A_{ij} = \frac{\exp\left(\frac{\mathbf{Q}_i \cdot \mathbf{K}_j^\top}{\sqrt{d}}\right) m_j}{\sum_k \exp\left(\frac{\mathbf{Q}_i \cdot \mathbf{K}_k^\top}{\sqrt{d}}\right) m_k}.
   \]
   This is equivalent to each key $\mathbf{K}_j$ and corresponding value $\mathbf{V}_j$ are duplicated $m_j$ times. 
   We discard the values corresponding to the mask, so the result of the attenion mechanism is the same. 
Thus, the attention mechanism with $\boldsymbol{\delta}$ scaling is mathematically equivalent to duplicating the keys and values proportionally to the number of times the mask appears.

\subsection{Token Length in MRM during Inference} To clarify the MRM's mechanism during training and inference, we include a more detailed explaination. One single masked token replaces all selected image tokens to mask during training. Eq.4 computes exactly same attention between   
\token\token\token\maskednumtoken{1}\maskednumtoken{1} and  \token\token\token\maskednumtoken{2} where the black box is the mask token the number inside represents the attention offset ($\delta$ in Eq.4). So in inference, we append \maskednumtoken{1} with $1$ (essentially no repeat) to make the token length same. \textit{Eg}:
\begin{center}
\vspace{-2mm}
Sample 1: \token\token\token\token \quad\quad
Sample 2: \token\token\maskednumtoken{1}\maskednumtoken{1}
\end{center}

\section{Performance}

\subsection{WebBody4M vs WebFace4M Comparison}

To assess the quality of the face image data within WebBody4M, we create WebBody-Facecrop by cropping face from the WebBody datset. And we compare its face recognition performance against WebFace4M~\cite{zhu2021webface260m}, a dedicated large-scale face recognition dataset. We train the same ViT-based model with AdaFace loss on both datasets. 
Tab.~\ref{tab:twodatasetscomparison}  presents the results on standard face recognition benchmarks (LFW, CPLFW, CFPFP, CALFW, and AGEDB). The model trained on WebBody4M achieves a slightly higher average accuracy (97.63\%) compared to that of WebFace4M (97.44\%). This indicates WebBody4M label is of comparable quality, even slightly exceeding WebFace4M label. 

\begin{table}[!t]
    \centering
    \scriptsize
    \begin{tabular}{ccccccc}
    \hline
         Dataset & Avg & LFW & CPLFW & CFPFP & CALFW & AGEDB \\ \hline
        WF4M & 97.44 & 99.80 & 94.97 & 98.94 & 96.03 & 97.48 \\
        \makecell{WB4M-\\Facecrop} & \textbf{97.63} & \textbf{99.82} & \textbf{95.12} & \textbf{99.19} & \textbf{96.07} & \textbf{97.97 }\\ \hline
    \end{tabular}
    \caption{Performance Comparison between WebFace4M and WebBody4M in the Face Recognition Task.}
\label{tab:twodatasetscomparison}
\end{table}

\subsection{Fusion Performance}

While SapiensID inherently handles both face and body information within a single model, a common alternative approach involves training separate face and body recognition models and fusing their outputs. We compare SapiensID's performance with such multi-modal fusion methods. We consider a baseline where a body model (CAL~\cite{gu2022clothes}) is trained on either PRCC or LTCC, and a face model (ViT-Base~\cite{kim2022adaface}) is trained on WebFace4M. We then fuse the similarity scores of these two dedicated face and body models using three common fusion strategies: Max Fusion, Min-Max Normalization Fusion, and Mean Fusion. Tab.~\ref{tab:fusion} presents the performance.

As shown in the table, even the best fusion strategy (Mean Fusion) achieves an average mAP of 49.99\%, lower than SapiensID's 52.87\%. Fusion is more helpful in PRCC but not much in LTCC with an increase in Top1 and a decrease in mAP. This result highlights the advantage of SapiensID's unified architecture, which learns to integrate face and body information more effectively than post-hoc fusion methods. Fusion methods treat each modality independently, potentially missing valuable contextual information that arises from their combined analysis.

\begin{table}[t]
    \centering
    \scriptsize

\begin{tabular}{l|ccc|ccc}
\hline
\textbf{} & \multirow{2}{*}{AVG} & \multicolumn{2}{c|}{LTCC CC} & \multicolumn{2}{c}{PRCC CC} \\
\textbf{} &  & Top1 & mAP & Top1 & mAP \\\hline
Body & 42.04 & 38.01 & \textbf{18.84} & 55.69 & 55.63 \\
Face & 36.56 & 17.60 & 4.91 & 72.62 & 51.10 \\\hline
Fused-Max & 42.93 & 39.80 & 13.25 & 61.22 & 57.45 \\
Fused Min-Max & 49.92 & 39.80 & 12.95 & 79.00 & 67.93 \\
Fused-Mean & 49.99 & 39.80 & 12.82 & \textbf{79.48} & 67.85 \\\hline
SapiensID & \textbf{52.87} & \textbf{42.35} & 17.79 & 78.75 & \textbf{72.60} \\
\hline
\end{tabular}

    \caption{Performance table of score fusion (Body and Face). }
    \label{tab:fusion}
\end{table}

\begin{table}[t]
\centering
\scriptsize
\begin{tabular}{cc|cc}
\hline
\multicolumn{2}{r|}{Method}   & KPR~\cite{somers2025keypoint} + SOLDIER & SapiensID \\
\multicolumn{2}{r|}{Training Data}   & LUPerson4M + OccludedReID & WebBody4M \\ \hline
\multirow{2}{*}{OccludedReID} & top1 & 84.80 & \textbf{87.30} \\
 & mAP & \textbf{82.60} & 75.57 \\\hline
\multirow{2}{*}{LTCC General} & top1 & 68.15 & \textbf{74.24} \\ 
 & mAP & 32.42 & \textbf{36.88} \\ \hdashline
\multirow{2}{*}{LTCC CC}  & top1 & 21.17 & \textbf{42.60} \\ 
 & mAP & 10.19 & \textbf{17.39} \\ \hline
\end{tabular}
\caption{Generalization performance comparison under occlusion. SapiensID demonstrates superior generalization to unseen datasets (LTCC) compared to KPR+SOLDIER.}
\label{tab:occlusion_generalization}
\end{table}

\begin{table*}[t]
    \centering
    \scriptsize
    \begin{minipage}{0.49\textwidth}
        \centering
        \begin{tabular}{cccccc}
        \hline
         &  & \multicolumn{2}{c}{LTCC CC} & \multicolumn{2}{c}{PRCC CC} \\ 
         &  & Top1 & mAP & Top1 & mAP \\ \hline
        1 & None & 0.00 & 3.56 & 1.47 & 4.28 \\ 
        2 & 1+Nose & 25.77 & 5.78 & 27.21 & 21.04 \\ 
        3 & 2+Eye & 30.61 & 8.87 & 63.87 & 55.17 \\ 
        4 & 3+Mouth & 38.01 & 11.81 & 73.36 & 65.05 \\ 
        5 & 4+Ear & 39.80 & 14.05 & 77.65 & 70.45 \\ 
        6 & 5+Shoulder & 41.84 & 15.82 & 79.73 & 73.14 \\ 
        7 & 6+Elbow & 41.07 & 16.64 & \textbf{80.55} & \textbf{73.54} \\ 
        8 & 7+Wrist & 41.07 & 17.16 & 79.34 & 73.16 \\ 
        9 & 8+Hip & 40.56 & 17.50 & 79.99 & 73.38 \\ 
        10 & 9+Knee & 42.35 & 17.73 & 79.00 & 72.88 \\ 
        11 & 10+Ankle (full) & \textbf{42.35} & \textbf{17.79} & 78.75 & 72.60 \\ \hline
        \end{tabular}
        \\(a) top-down

    \end{minipage}
    \hfill
\begin{minipage}{0.49\textwidth}
    \centering
    \begin{tabular}{cccccc}
    \hline
     &  & \multicolumn{2}{c}{LTCC CC} & \multicolumn{2}{c}{PRCC CC} \\ 
     &  & Top1 & mAP & Top1 & mAP \\ \hline
    1 & None & 0.00 & 3.56 & 1.47 & 4.28 \\ 
    2 & 1+Ankle & 27.04 & 7.37 & 45.05 & 35.32 \\ 
    3 & 2+Knee & 32.14 & 9.55 & 55.12 & 44.97 \\ 
    4 & 3+Hip & 35.71 & 12.34 & 66.07 & 55.04 \\ 
    5 & 4+Wrist & 37.24 & 13.83 & 67.63 & 58.43 \\ 
    6 & 5+Elbow & 40.05 & 15.72 & 69.57 & 62.61 \\ 
    7 & 6+Shoulder & 41.33 & 16.87 & 73.84 & 67.80 \\ 
    8 & 7+Ear & 41.58 & 17.61 & 76.21 & 70.62 \\ 
    9 & 8+Mouth & 41.58 & 17.95 & 78.18 & 72.63 \\ 
    10 & 9+Eye & 41.58 &\textbf{ 17.80} & \textbf{79.23} & \textbf{72.92} \\ 
    11 & 10+Nose (Full) & \textbf{42.35} & 17.79 & 78.75 & 72.60 \\ \hline
    \end{tabular}\\
    (b) bottom-up
\end{minipage}
\caption{Comparison of feature erasing performance. (a) shows the performance as we progressively introduce features from Nose to Ankle (top-down approach). (b) demonstrates the performance when adding features from Ankle to Nose (bottom-up approach). Results are evaluated on LTCC and PRCC Cloth Changing (CC) protocol.}
\label{tab:featuremask}

\end{table*}

\begin{table*}[!t]
    \centering
    \scriptsize
    \begin{minipage}{0.49\textwidth}
        \centering
        \begin{tabular}{cccccc}
        \hline
         &  & \multicolumn{2}{c}{LTCC CC} & \multicolumn{2}{c}{PRCC CC} \\ 
         &  & Top1 & mAP & Top1 & mAP \\ \hline
        1 & None & 2.30 & 1.89 & 12.67 & 4.78 \\ 
        2 & 1+Top1 & 5.10 & 2.61 & 78.04 & 67.29 \\ 
        3 & 2+Top2 & 27.04 & 11.88 & 79.25 & 70.53 \\ 
        4 & 3+Top3 & 29.34 & 13.20 & 78.35 & 69.85 \\ 
        5 & 4+Top4 & 33.67 & 13.88 & 77.82 & 69.55 \\ 
        6 & 5+Top5 & 37.24 & 14.65 & 76.97 & 69.28 \\ 
        7 & 6+Top6 & 36.48 & 15.49 & 78.55 & 70.39 \\ 
        8 & 7+Top7 & 41.07 & 16.63 & \textbf{80.07} & \textbf{71.52} \\ 
        9 & Full & \textbf{42.35} & \textbf{17.79} & 78.75 & 72.60 \\ \hline
        \end{tabular}\\
        (a) top-add
        \label{tab:partfromtop}
    \end{minipage}
    \hfill
    \begin{minipage}{0.49\textwidth}
        \centering
        \begin{tabular}{cccccc}
        \hline
         &  & \multicolumn{2}{c}{LTCC CC} & \multicolumn{2}{c}{PRCC CC} \\ 
         &  & Top1 & mAP & Top1 & mAP \\ \hline
        1 & None & 2.30 & 1.87 & 12.50 & 4.78 \\ 
        2 & 1+Bottom1 & 2.81 & 2.26 & 24.56 & 10.89 \\ 
        3 & 2+Bottom2 & 6.12 & 3.08 & 31.22 & 16.94 \\ 
        4 & 3+Bottom3 & 5.87 & 3.62 & 33.78 & 20.65 \\ 
        5 & 4+Bottom4 & 10.20 & 4.26 & 33.08 & 24.59 \\ 
        6 & 5+Bottom5 & 12.50 & 5.33 & 22.10 & 21.31 \\ 
        7 & 6+Bottom6 & 16.07 & 6.48 & 24.47 & 24.80 \\ 
        8 & 7+Bottom7 & 35.46 & 13.20 & 29.07 & 28.63 \\ 
        9 & Full & \textbf{42.35} & \textbf{17.79} & \textbf{78.75} & \textbf{72.60 }\\ \hline
        \end{tabular}\\
        (b) bottom-add
        
    \end{minipage}
    \caption{Impact of progressively adding visible parts from the (a) top and from the (b) bottom. In contrast to Tab.~\ref{tab:featuremask} which measures the performance with the intermediate features zeroed out, here the actual input image is masked out.}
     
    \label{tab:imageerase}
\end{table*}

\subsection{Occluded ReID}

Occlusions pose a significant challenge for robust human recognition. While specialized methods can be effective within their training domain, generalization to unseen scenarios is crucial for real-world deployment. We compare SapiensID's performance with KPR \cite{somers2025keypoint} combined with SOLDIER, a state-of-the-art occlusion handling method, to evaluate their respective generalization capabilities. KPR+SOLDIER is trained on a combination of LUPerson4M and the OccludedReID \cite{zhuo2018occluded} dataset, while SapiensID is trained on our WebBody4M dataset without any OccludedReID data.

Tab.~\ref{tab:occlusion_generalization} presents the results on OccludedReID and the LTCC dataset (both General and Clothing Change protocols).  KPR+SOLDIER and SapiensID similar performance on OccludedReID, SapiensID demonstrates significantly better generalization performance. On LTCC, SapiensID substantially outperforms KPR+SOLDIER across both protocols, highlighting the limitations of specialized training. This underscores the importance of training on diverse datasets like WebBody4M to achieve robust generalization in real-world human recognition. SapiensID, by learning from a wide range of poses, viewpoints, and clothing styles, is more adaptable and effective in unseen scenarios.

\subsection{Impact of Body Part Features}
\label{supp:part}

We investigate the relative importance of different body parts in human recognition by conducting an ablation study on the Semantic Attention Head (SAH). Starting from part features ($\mathbf{O}_{part}^i$ in Eq.~\ref{eq:sah_attention}) multiplied by zero, we progressively undo masking, either from nose-to-ankles (top-down) or ankles-to-nose (bottom-up). We evaluate performance on LTCC (Clothing Change protocol) and PRCC (Clothing Change protocol). Results are presented side-by-side in Tab.~\ref{tab:featuremask}. The top-down approach generally yields faster performance gains than bottom-up, suggesting that upper-body features contribute more significantly to recognition. 

Interestingly, ankle features alone appear more discriminative than nose features alone. However, this counter-intuitive finding does not imply that ankles are inherently more informative than noses for person identification. We hypothesize that this observation arises because each part feature within SAH is not solely derived from the corresponding body part. Due to the preceding ViT backbone's attention mechanism, each part feature incorporates information from other body regions. Therefore, the presented results reflect the discriminative power of a part plus peripheral information from other parts, rather than the isolated contribution of each part.

A more accurate assessment of a part's individual discriminative ability would involve manipulating the input image directly, such as by occluding specific body parts. This approach, which isolates the impact of each part, is explored in the following section.

\subsection{Impact of Actual Image Erased}

\begin{figure}[t]
    \centering
    \includegraphics[width=1.0\linewidth]{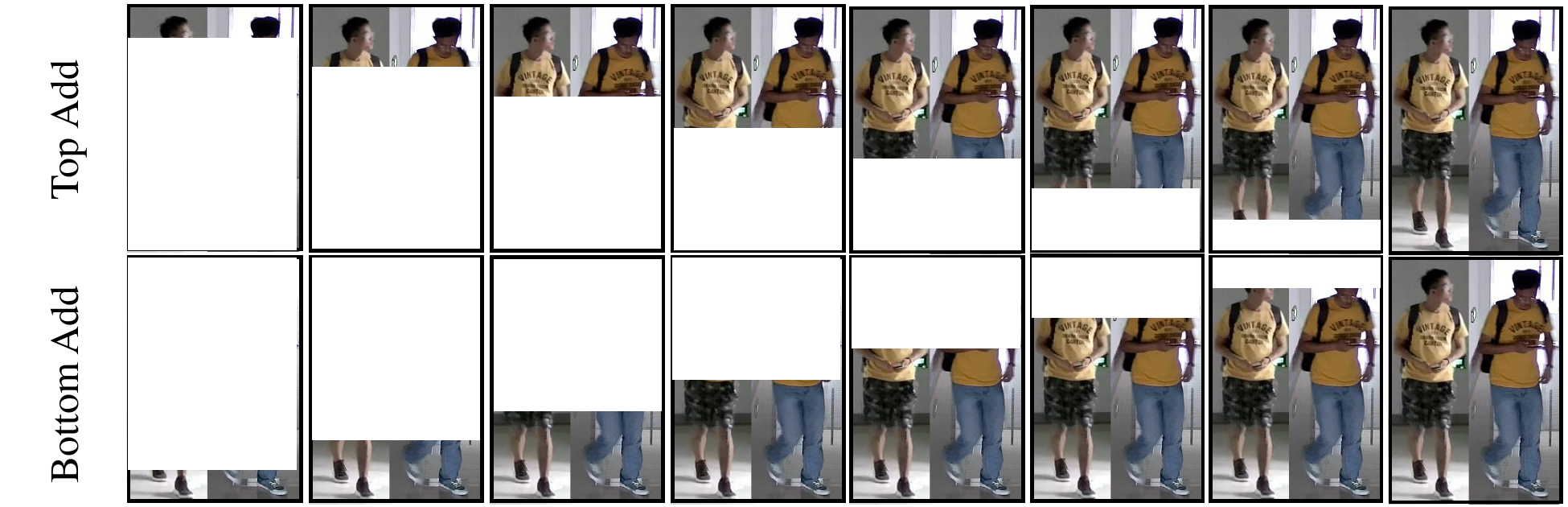}
    \caption{Illustration of how Images are erased from top to bottom or bottom to top. }
    \label{fig:eraseimage}
\end{figure}

To isolate the contribution of each body region, we conduct a second ablation study where we progressively erase sections of the input image, either top-down or bottom-up, as illustrated in Fig.~\ref{fig:eraseimage}. We erase equal-sized horizontal strips, starting with a single strip and progressively adding more until the whole image is erased (represented as "None" in the tables). The "Full" row represents the baseline performance with the complete image. Results are presented in Tab.~\ref{tab:imageerase}.

The direct manipulation of the image confirms the importance of upper body regions. On both datasets, removing the top portion of the image drastically reduces performance. It comes as a surprise that PRCC can achieve a very good performance with only 1 top strip of image. But for LTCC, the lower parts are necessary to obtain a good performance. This indicates that different datasets exhibit different characteristics that can be exploited for conducting ReID.

\subsection{Sensitivity to Pose Estimation }
To understand the sensitivity of SapiensID to the pose estimation, we compare OpenPose~\cite{openpose}, and YoloV8~\cite{yolov8_ultralytics} and add Gaussian noise ($\sigma=0.01$). Tab (a) shows minimal impact from detector choice, but systematic keypoint errors reduce performance. Contrarily, in (b) we show how 5\% zoom degrades CLIP3DReID, while SapiensID remains robust, making it the first ReID model robust to input extrinsics.
\begin{figure}[h!]
    \centering
    \begin{minipage}{0.57\linewidth} 
        \centering
        \scriptsize
            \renewcommand{\arraystretch}{0.75}

        \begin{tabular}{c|c|c} \hline    
        \multirow{2}{*}{ Keypoint Predictor}  & \multicolumn{2}{c}{Whole Body ReID} \\
                  & Short & Long \\ \hline
        Open Pose& 66.30&  73.05
        \\ 
        Yolo V8& 65.62&  72.76
        \\ 
        Open Pose + $\epsilon$& 56.08&  65.72\\ \hline
        \end{tabular}\\
        (a) SapiensID with keypoint changes 
        
        \vspace{0.3em} 

        \begin{tabular}{c|c|c} \hline    
        \multirow{2}{*}{ Extrinsic Change}  & \multicolumn{2}{c}{LTCC (CC)} \\
                  & Original & Zoom 5\% \\ \hline
        CLIP3DReID[\textcolor{cvprblue}{44}] & 41.84 & \textcolor{red}{31.88} \\ 
        SapiensID & 42.35 & \textcolor{black}{41.58} 
        \\ \hline
        \end{tabular}\\
        (b) Different camera extrinsics
    \end{minipage}%
    \hspace{0.5em} 
    \begin{minipage}{0.39\linewidth} 
        \centering
    \includegraphics[width=\linewidth]{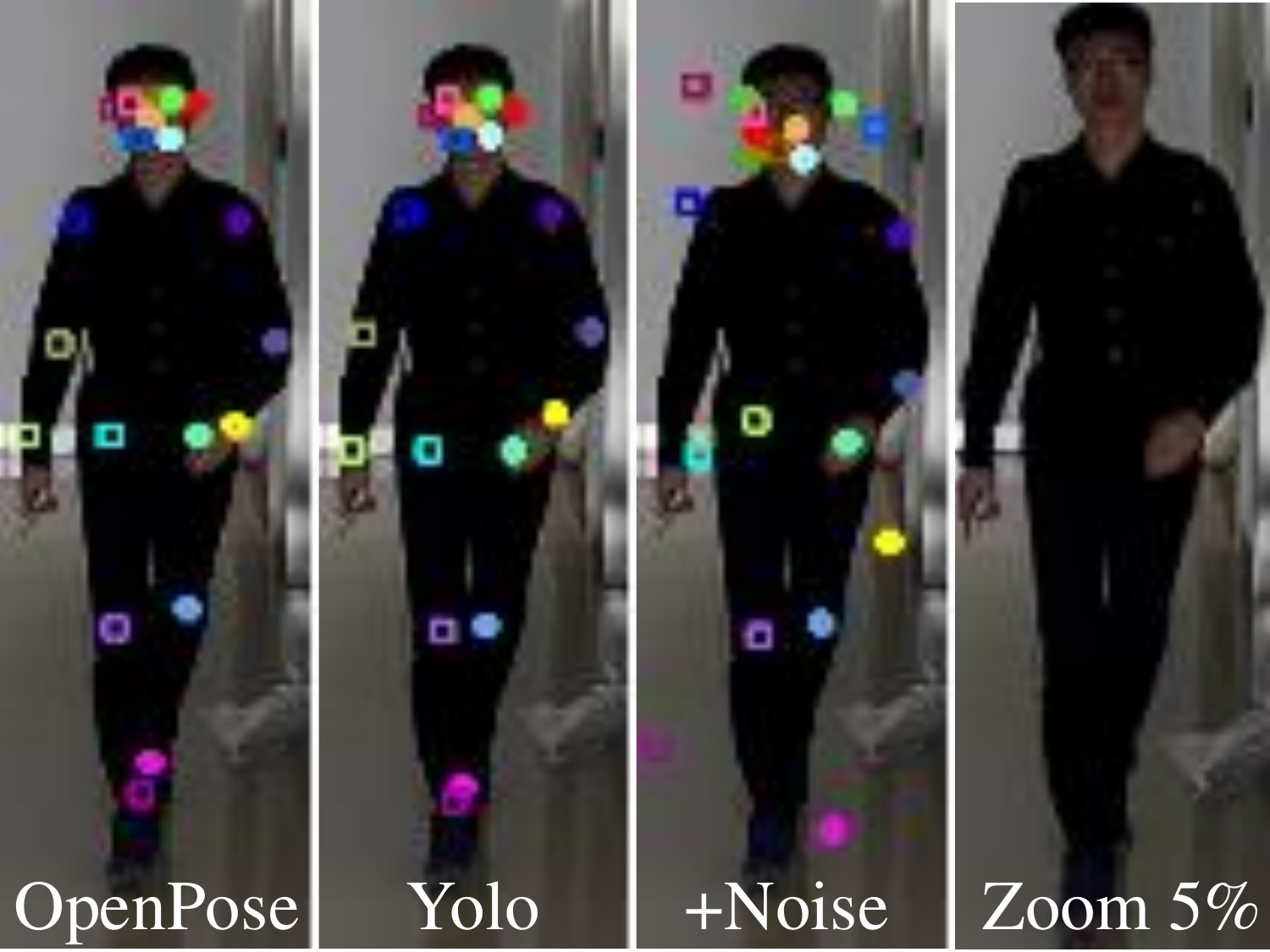}
    \scriptsize
    (c) Example visualization
    \end{minipage}
        \label{fig:kpsensitive}
\end{figure}

\subsection{(Ablation on Model Size}
We investigate the relationship between performance and the model and dataset size. In Tab.~\ref{tab:modelsize_rebut}, we include ViT size variation (small vs base). The trend shows that the larger model has higher performance. We also created WebBody12M, in addition to 4M and the dataset increase further improves the performance. 
\begin{table}[h!]
\centering
\scriptsize

\renewcommand{\arraystretch}{0.92}
\begin{tabular}{c |c |c |c|c|c|c} \hline    
SapiensID & Dataset& LTCC& CCDA& Celeb& LFW& AGEDB
\\ \hline 
Small& WB4M&  71.40&  57.04& 91.29& 99.67& 96.58
\\ 
Base& WB4M&  74.24&  61.84& 92.77& 99.77& 97.18
\\ 
Base& WB12M&  \textbf{75.66}&  \textbf{66.80}& \textbf{94.01}&\textbf{ 99.85}& \textbf{98.02}\\ \hline

\end{tabular}
\caption{SapiensID backbone and dataset size Variation.  }
\label{tab:modelsize_rebut}
\end{table}

\subsection{FLOP Analysis}
In this subsection, we provide the FLOP analysis of SapiensID. The backbone model shares face model backbone (ViT-base). The major difference with ViT-base is the number of tokens. In inference, RetinaPatch produces $281$ tokens on average (vs. $196$ in ViT), increasing FLOPs from $24.69$G to $35.39$G. RetinaPatch ($0.45$G FLOPs) and Head ($1.09$G FLOPs, $336.65$M params) add minimal overhead. Similarity measure is cosine dist, same as ArcFace. 

\subsection{Role of Masked Recognition Model (MRM) }
In this subsection, we provide more ablation of MRM to showcase the importance of variable masking rate. Starting from simple ViT, we progressively add elements that comprises MRM. First we introduce token masking to handle varying token counts from RetinaPatch and improve training speed. 
Yet, simple masking significantly reduces performance due to discrepancies between training and testing samples. Thus, we propose variable rate masking (MRM), which restores performance to full-token training levels (see the table below, row 1 vs 3). All performance is measure without Retina Patch or Semantic Attention Pooling.   
\begin{table}[!h]
    \centering
    \scriptsize
    \renewcommand{\arraystretch}{0.92}
    \begin{tabular}{l|ccc}
    \hline
         \multirow{2}{*}{ Metric same as Tab.5 (main paper)} & \multirow{2}{*}{Face} & \multicolumn{2}{c}{Whole Body ReID} \\
        &  & Short & Long \\ \hline
        (1) ViT (Full Token) & 90.63& 56.17& 31.81
\\ 
        (1)  + Token Masking  (always remove 33\%) & 57.73& 49.23& 25.83
\\ 
        (1)  + MRM (variable remove rate) & 89.54& 55.56& 30.76\\\hline 
        (1)  + MRM  + Retina Patch &  92.93 & 59.16 & 46.95 \\\hline 
    \end{tabular}
\caption{Performance of ViT as measured in Tab.5 of main paper. MRM is needed to allow Retina}
\label{tab:mrmrole}
\end{table}

\subsection{Additional Face Recognition (FR) Perf.}
We include more face recognition performances to investigate the performance of SapiensID in more challenging face recognition scenarios. We include the performance measured in IJB-B~\cite{ijbb}, IJB-C~\cite{ijbb}, TinyFace~\cite{tinyface}. TinyFace measures the face recognition performance in low quality imageries. 
WebBody4M is actually rich in small faces due to whole body images. 
It results in better TinyFace performances (row 2,3) than WebFace4M. 
SapiensID, inherently a body ReID model works well on aligned faces is because RetinaPatch always focuses on the face region. 
\begin{table}[h!]
\centering
\scriptsize
\renewcommand{\arraystretch}{0.92}
\begin{tabular}{l |l|c |c |c |c} \hline    
 \multirow{2}{*}{ Aligned FR }  &\multirow{2}{*}{ Training Data} & IJBB & IJBC& \multicolumn{2}{c}{TinyFace}\\ 
    &   & \multicolumn{2}{c|}{TAR@FAR0.01\%}& R1& R5\\ \hline 
ViT-AdaFace & WebFace4M &  95.60 &  97.14 &  74.81 &  77.58
\\
 ViT-AdaFace & WB4M (Face crop) & \textbf{95.92} & \textbf{97.22} & 75.32 & 78.76
\\
SapiensID  &WB4M &  95.07 &  96.43 &  \textbf{75.97} &  \textbf{79.69}\\ \hline

\end{tabular}
\caption{Face Recognition Performance with ViT-Base. IJB,C measured in TAR@FAR=0.01\%. All input images are aligned.}
\label{tab:addface}
\end{table}

\subsection{IJB-S Evaluation}
A unified model is useful when matching cross modality imagery. In IJB-S~\cite{ijbs} evaluation Surv2Single protocol, probe surveillance videos are matched to close-up gallery face images. UAV2Book presents an even greater challenge, with drone-captured probe videos featuring smaller faces and high-pitch angles. 
In such case, facial regions are too small.
With a shared representation for both the whole body and face, the unified model (SapiensID) {\it opportunistically} captures more contextual cues, leading to improved matching, as shown below. Separate face or body models don't share the same representation space to conduct cross-modality matching. 
All models are finetuned on LQ BRAIR dataset.
\begin{table}[!h]
\renewcommand{\arraystretch}{0.92}
\centering
\scriptsize
\begin{tabular}{c|cc|cc|cc} 
\hline
 \multicolumn{1}{c|}{IJB-S Evaluation}& \multicolumn{2}{c|}{Input Type}&\multicolumn{2}{c|}{Surv2Single}& \multicolumn{2}{c}{UAV2Book}\\
Model & Probe&Gallery& R1 & R5 & R1 & R5 \\\hline

Body Models & Body  &Face &  \multicolumn{4}{c}{NA because raw gallery is face.} \\\hline
ViT-AdaFace & Face  &Face& 75.6 & 79.7 & 29.1 & 38.0 \\\hline
\multirow{2}{*}{SapiensID } & Face  &Face& \textbf{75.8} & \textbf{80.0} & 31.6 & 44.3 \\
& \textbf{Body}&Face& 72.6 & 77.9 & \textbf{39.2} & \textbf{49.4} \\\hline

\end{tabular}
\caption{Performance in IJB-S Evaluation Dataset.}
\end{table}

\subsection{Unaligend Face Recognition}
We also show unaligned IJB-B/C results to see the face recognition performance without alignment. A dedicated FR model is better in aligned, but SapiensID has less performance drop in unaligned settings. 
\begin{table}[h!]
\centering
\renewcommand{\arraystretch}{0.92}
\scriptsize
\begin{tabular}{c|ccll}
\hline
 \multirow{2}{*}{Metric TAR@FAR=0.01\%}  & \multicolumn{2}{c}{Unaligned} & \multicolumn{2}{c}{Aligned}\\ 
       & IJB-B& IJB-C& IJB-B&IJB-C\\ \hline
ViT-AdaFace & 93.26& 94.97& \textbf{95.60}&\textbf{97.14}
\\
SapiensID & \textbf{94.30}& \textbf{96.05}& 95.07&96.43\\ 
\hline

\end{tabular}
\label{tab:ibbbc}
\end{table}

\section{Visualization}

\begin{figure}[t]
    \centering
    \includegraphics[width=\linewidth]{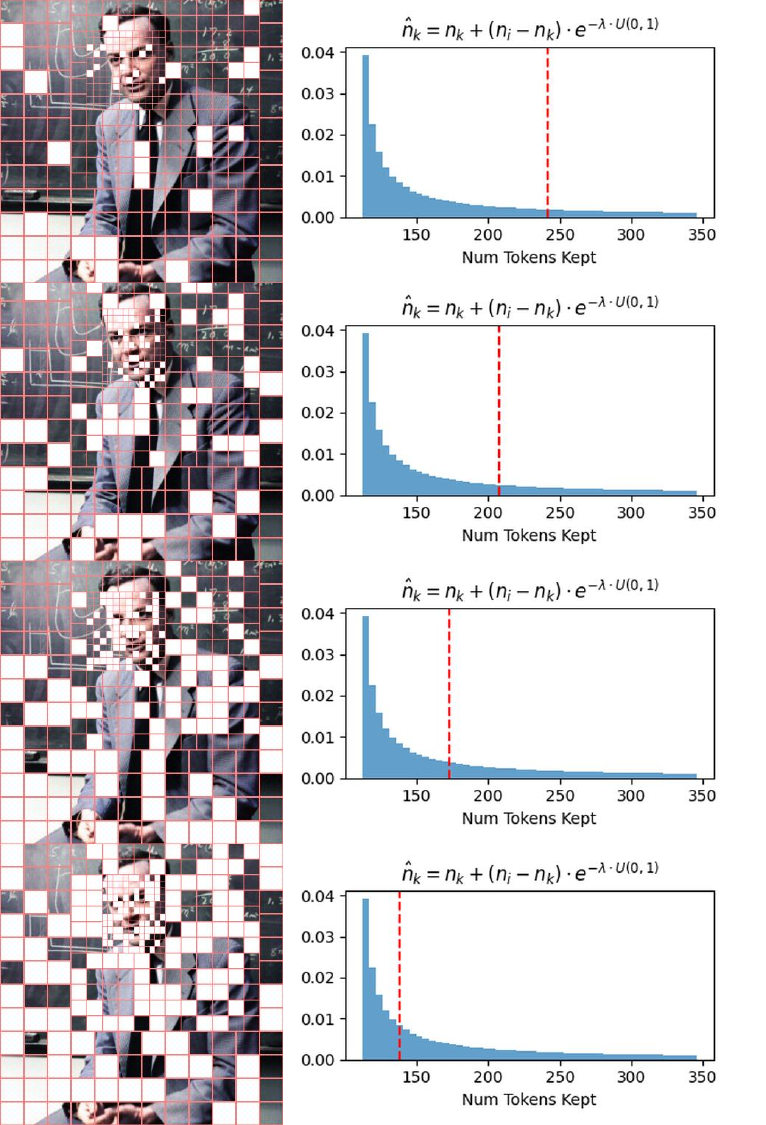}
    \caption{Illustration of the masked image and the sampling distribution of the number of tokens to keep $\hat{n}_k$. The red vertical line shows where the sampling took place for the right image. From top to bottom, less samples are kept (more masking).}
    \label{fig:sample}
\end{figure}

\subsection{Token Length Sampling Distribution}

In Masked Recognition Model (MRM), we propose an adaptive token sampling strategy during training to enhance the robustness and generalization of our masked recognition model. Fig.~\ref{fig:sample} illustrates the sampling distribution and its effect on the input image. The number of tokens to keep, $\hat{n}_k$, is determined by Eqn.~\ref{eq:sample}:
$$\hat{n}_k = n_k + (n_i - n_k) \cdot e^{-\lambda \cdot U(0, 1)},$$
where $n_i$ is the maximum possible number of tokens (432 in our case, with 3 ROIs of 12x12 patches each), $n_k$ is the minimum number of tokens to keep, $U(0,1)$ is a uniform random variable, and $\lambda$ controls the decay rate (set to 4).

This sampling strategy allows us to retain between 26\% and 80\% of the tokens (112 to 345 tokens), with an average of 166 tokens per batch. As depicted in Fig.~\ref{fig:sample}, heavy masking can significantly distort the input image. Fixing the masking rate to such high levels could introduce a distribution shift between training and testing (where all tokens are used), causing a performance drop. Our adaptive sampling mitigates this issue by exposing the model to a variety of masking ratios, encouraging it to learn robust representations that generalize well to full token input during inference.

One thing to note is that the sampling of $\hat{n}_k$ happens per batch. And when a larger $\hat{n}_k$ is sampled per batch, we reduce the batch size accordingly for the given GPU memory (See Sec.~\ref{sec:mrb} for more details).

\subsection{WebBody4M Dataset Body Parts Visibility}

WebBody4M dataset encompasses a wide range of human poses and viewpoints, resulting in varying visibility of body keypoints. Tab.~\ref{tab:keypoints} presents the percentage of images in which each keypoint (left and right sides) is visible. As expected, keypoints in the upper body, such as eyes and shoulders, exhibit high visibility rates (over 74\% and 88\% respectively). Visibility decreases progressively down the body, with elbows and wrists around 50\%, hips around 45\%, and knees and ankles below 24\% and 17\% respectively. This distribution reflects the natural tendency for upper body parts to be more frequently visible in unconstrained images, as lower body parts are often occluded by clothing, objects, or the image frame itself. This distribution also helps explain why upper body parts provide greater discriminative power for person ReID in our earlier analysis (Supp~\ref{supp:part}).

\begin{table}
\centering
\scriptsize
\begin{tabular}{ccc}
\hline
Visibility & Left (\%) & Right (\%) \\
\hline

Eye & 93.49 & 93.59 \\
Ear & 76.87 & 74.48 \\
Shoulder & 88.15 & 90.04 \\
Elbow & 53.76 & 53.80 \\
Wrist & 49.98 & 50.35 \\
Hip & 45.68 & 45.70 \\
Knee & 23.92 & 23.95 \\
Ankle & 16.98 & 17.00 \\
\hline
\end{tabular}
\caption{Keypoint Visibility in WebBody Dataset.}
\label{tab:keypoints}
\end{table}

\begin{figure*}[t!]
    \centering
    \includegraphics[width=1.0\linewidth]{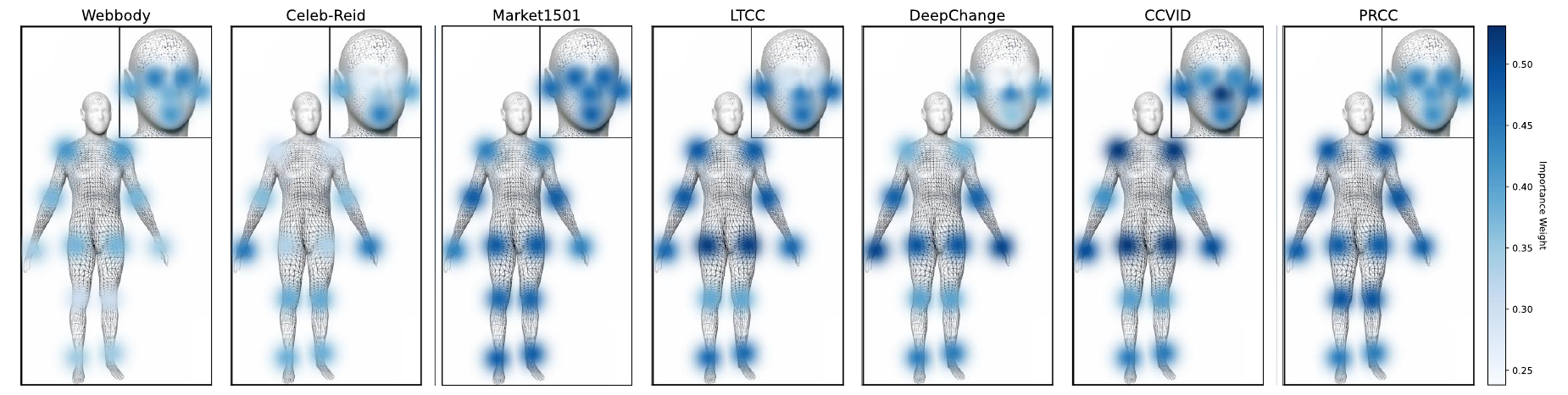}
    \caption{Comparison of learned part weights across seven datasets. Left and right sides are averaged together before visualization. }
    \label{fig:dataweight}
\end{figure*}

\subsection{Visualization of Part Weights}

To facilitate effective learning from a mixture of short-term and long-term ReID datasets, we hypothesize that it would be helpful to add learnable weights that modulate the importance of individual part features within the Semantic Attention Head (SAH). Our conjecture is the discriminative characteristics of body parts can vary significantly depending on whether clothing remains constant or varying in the training dataset.

Fig.~\ref{fig:dataweight} visualizes the learned weights (Eqn.~\ref{eq:weights}) for WebBody4M and several additional whole-body ReID datasets. WebBody4M, primarily composed of web-collected images, exhibits a higher emphasis on facial features compared to lower body parts. This is expected, as the WebBody4M was collected largely based on facial similarity.

In contrast to WebBody4M, auxiliary datasets like Market1501, LTCC, and PRCC, which feature many images with consistent clothing (e.g., 1-3 outfits across 20-30 images per person), show increased emphasis on body features for recognition. This highlights the importance of body shape, pose, and clothing appearance as discriminative cues when attire remains relatively constant. However, Celeb-ReID, similar to WebBody4M, primarily contains images with clothing changes across captures. Consequently, Celeb-ReID exhibits a similar weighting pattern, with less emphasis on body features and a relatively higher focus on other cues, likely emphasizing facial features.

To validate the hypothesis, we conducted an ablation study to evaluate the impact of training with learnable weights. Tab.~\ref{tab:ablation_noweight} presents a comparison between SapiensID and SapiensID without the learnable weights. In the latter, all aspects remain the same except that the learnable weights are removed during training.

From the results, it is evident that the inclusion of learnable weights does not yield a significant overall improvement. Instead, it shows a specific enhancement in long-term ReID performance, possibly because WebBody4M's learning was not hindered by the influence of short-term datasets with same clothings. However, for short-term datasets, the addition of weights does not result in performance gains. This suggests that while the weighting mechanism provides insights into dataset-specific learning behaviors, it is not a definitive factor for achieving better ReID performance.

In conclusion, while the introduction of learnable weights is interesting for analytical purposes, we want to let the readers clearly know that it is not a deciding factor for learning universal representation that works for both short-term and long-term ReID. Future research could explore alternative methods that better balance the learning from diverse dataset characteristics without negatively impacting specific subsets.

\begin{table}[t!]
\centering
\scriptsize
\begin{tabular}{l|c|ccc}
\hline
         & \multirow{2}{*}{All} & \multirow{2}{*}{Face} & \multicolumn{2}{c}{Whole Body ReID} \\
         &                      &                      & Short         & Long         \\ \hline
SapiensID & \textbf{78.67}               & \textbf{96.66}               & 73.05         & \textbf{66.30}        \\ 
\multicolumn{1}{l|}{\makecell[l]{SapiensID-Weight}} 
         & 78.59         & \textbf{96.66}                & \textbf{75.72}         & 63.39        \\ \hline
\end{tabular}
\caption{Performance comparison of SapiensID and SapiensID without weight masking during training across different metrics.}
\label{tab:ablation_noweight}
\end{table}

\begin{figure*}[t]
    \centering
    \includegraphics[width=1.0\linewidth]{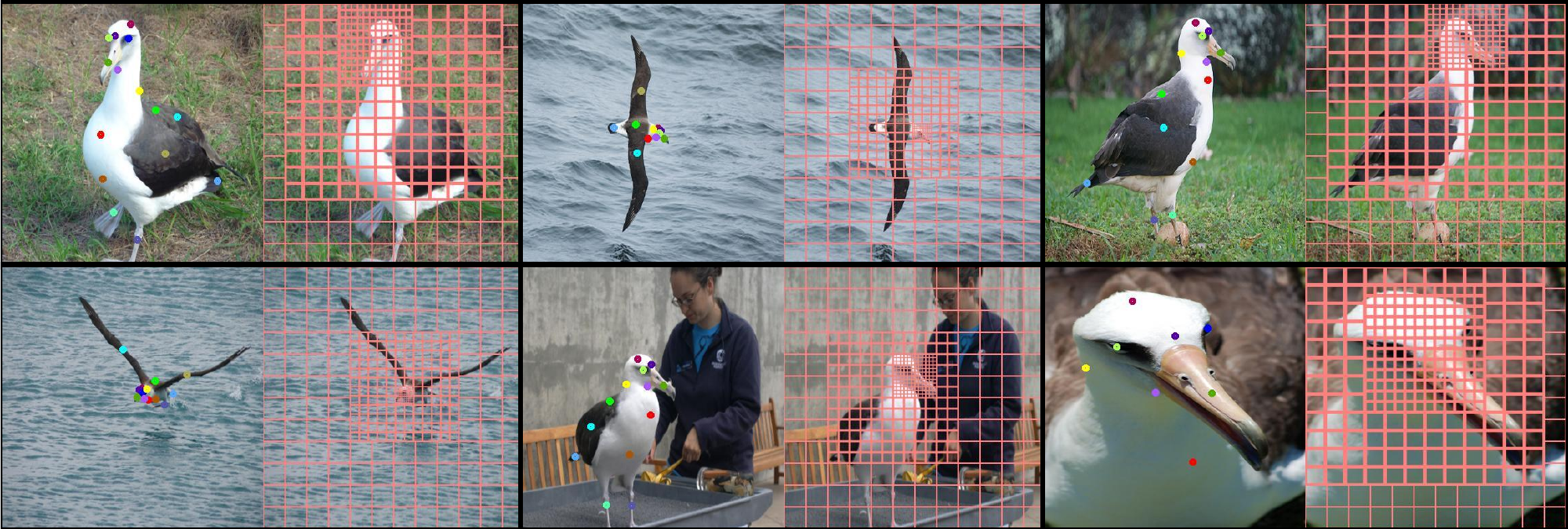}
    \caption{Keypoint visualization (left) and corresponding Retina Patch results (right) for images from the CUB dataset.}
    \label{fig:cub}
\end{figure*}

\subsection{SAH Visualization}

\begin{figure*}[t!]
    \centering
    \includegraphics[width=1.0\linewidth]{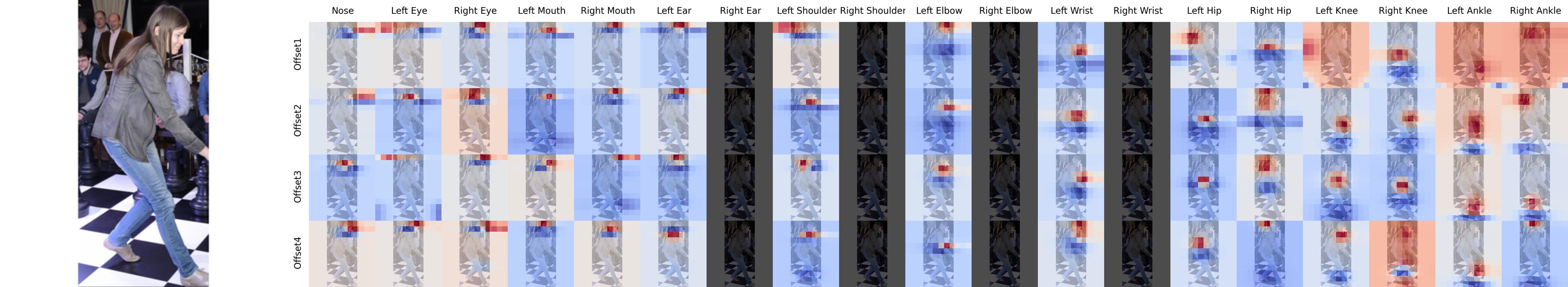}\\[2mm]
    \includegraphics[width=1.0\linewidth]{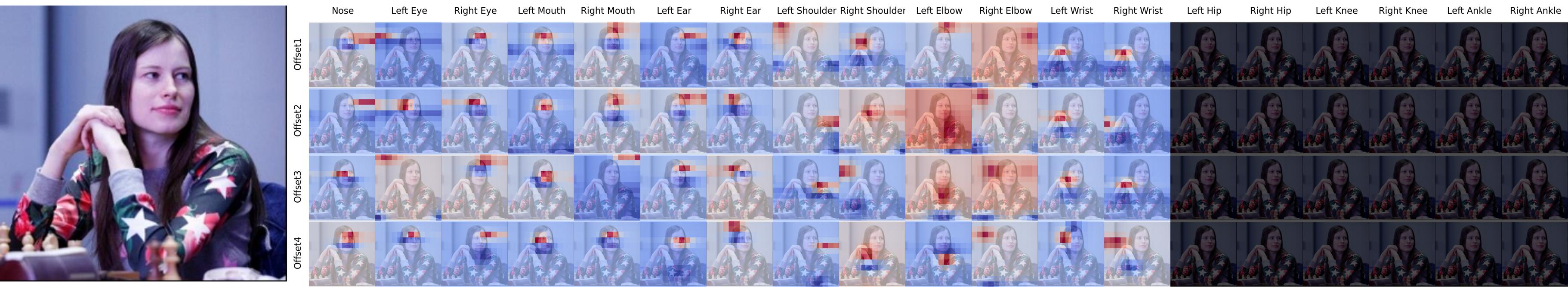}\\[2mm]
    \includegraphics[width=1.0\linewidth]{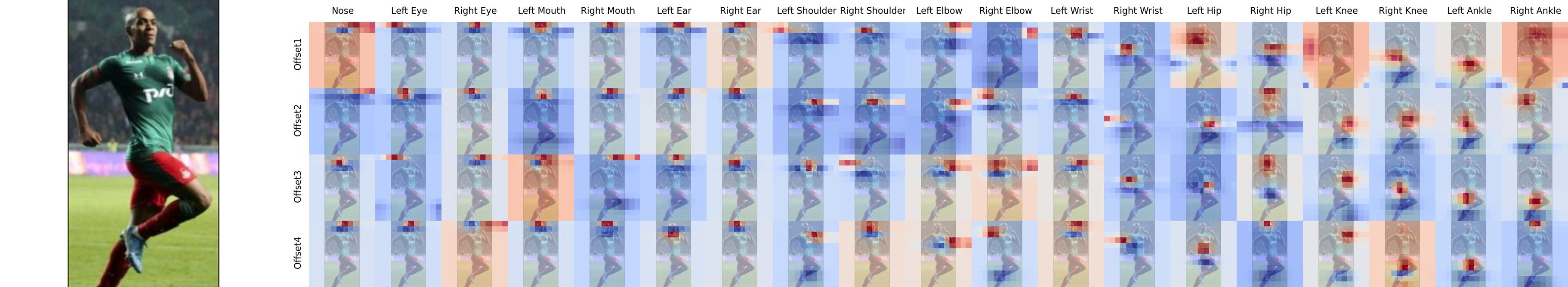}\\[2mm]
    \includegraphics[width=1.0\linewidth]{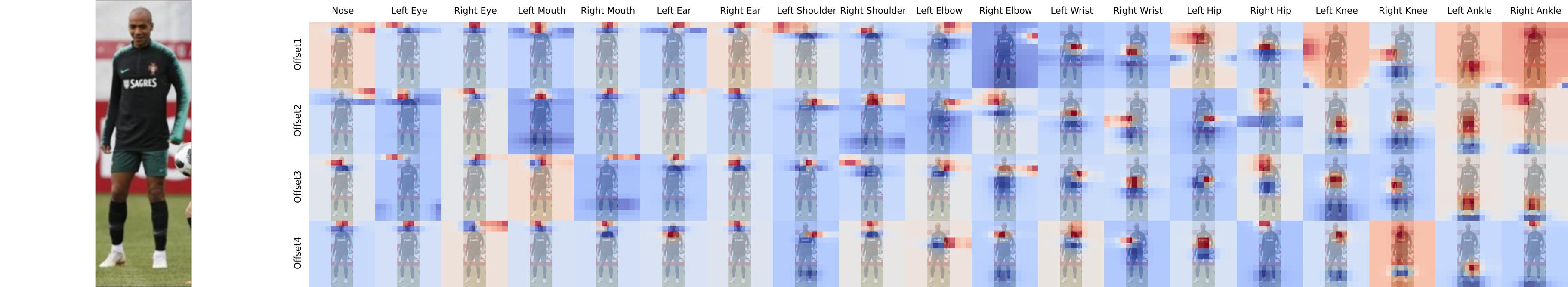}\\[2mm]
    \includegraphics[width=1.0\linewidth]{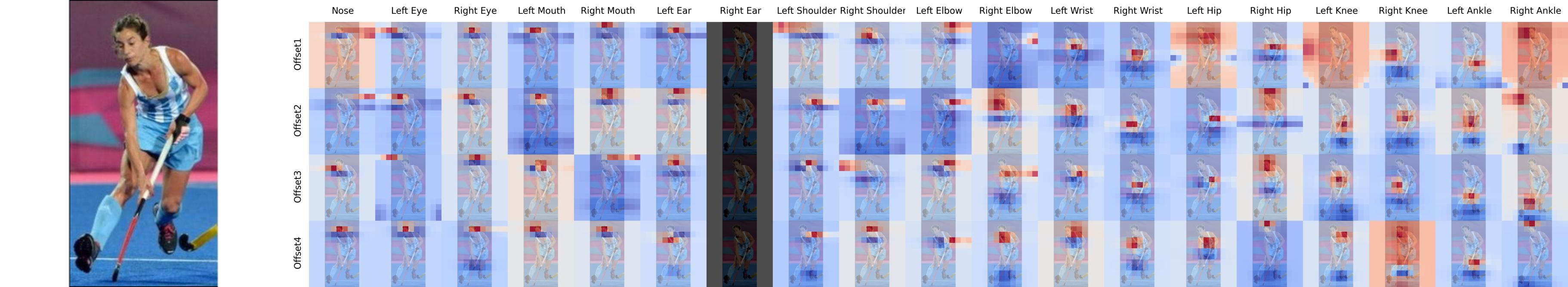}
    \caption{Visualization of attention maps in the Semantic Attention Head (SAH). Regions with higher attention values are highlighted in red, while regions with lower attention values are shown in blue. Blacked-out areas represent parts of the images without visible keypoints. The visualizations provides how SAH allows learning both varied size and offsets based on a set of keypoints. }
    \label{fig:attention_images}
\end{figure*}

The Semantic Attention Head (SAH) plays a crucial role in SapiensID by generating pose-invariant features. To understand how SAH behaves after training, we visualize its attention maps in Fig.~\ref{fig:attention_images}. To be specific, we visualize the following. Let  $\mathbf{Q}_{kp}^i = \text{GridSample}(\text{PE}, \text{kp}^i) + \mathbf{B}$ be the semantic query embedding for $i$-th image created by sampling from the fixed 2D position embeddings (PE) at the 19 keypoint locations. The dimension is $\mathbf{Q}_{kp}^i \in \mathbb{R}^{nk\times C}$, where $k=19$ and $n=4$ because it is repeated $4$ times to learn 4 different offsets. In SAH, we perform attention with  $\mathbf{Q}_{kp}^i$ and PE by 
\begin{equation}
\mathbf{O}_{\text{part}}^i = \text{softmax}\left(\frac{\mathbf{W}_q\mathbf{Q} \mathbf{W}_k\mathbf{K}^\top}{\sqrt{d}}\right) \mathbf{W}_v \mathbf{V}.
\end{equation}
In our visualization, we are showing 
$$\text{softmax}\left(\frac{\mathbf{W}_q\mathbf{Q} \mathbf{W}_k\mathbf{K}^\top}{\sqrt{d}}\right),$$
for each keypoint and each offset. We have $nk$ attention maps as shown by the visualization. 

For each input image, we show each row corresponds to a different offset. There are 4 rows because we learn $n=4$ offsets for each of $19$ keypoints. Offest refers to $\mathbf{B}\in \mathbb{R}^{nk\times C}$ in Eqn.~\ref{eq:sah}. Offset bias allows the keypoints to move slightly from its original position.
Each column correspond to different keypoints used by SAH (e.g., nose, left right shoulder, \textit{etc}). As the visualization shows, the learned attenion maps are not limited to the keypoint location but also move around the keypoints and vary in size.

\section{Potential Application of Retina Patch}

While SapiensID focuses on human recognition, the Retina Patch (RP) mechanism has broader applicability to other domains. Figure \ref{fig:cub} demonstrates its potential for fine-grained visual recognition, using the CUB birds dataset as an example. This dataset provides semantic keypoints, enabling the definition of meaningful regions of interest (ROIs) for RP. We define two ROIs: "head" (beak, forehead, crown, left eye, right eye, throat) and "body" (back, belly, breast, nape, left wing, right wing) excluding tail, left leg and right leg.

The figure showcases multiple bird images processed with RP, illustrating its ability to handle variations in bird size and head size. By dynamically allocating more patches to these regions, RP ensures consistent representation of crucial features, regardless of their scale within the image. Though we do not know whether the performance of CUB bird classification will be improved with RP, we want to suggest that RP could be beneficial for general recognition tasks where image naturally contains large pose and scale variation. Future work could explore the integration of RP into models for more broad set of datasets to quantitatively evaluate its benefits.

\section{Limitations}

While SapiensID demonstrates promising results for human recognition, its reliance on predefined Regions of Interest (ROIs) introduces certain limitations. The effectiveness of the Retina Patch mechanism hinges on the ability to define meaningful ROIs that capture discriminative features. This approach works well for humans, who share a consistent body topology and where keypoints like the face, torso, and limbs provide valuable cues for recognition.

However, this reliance on ROIs poses challenges when dealing with objects or entities that lack a consistent or well-defined structure. For instance, applying SapiensID to amorphous objects, scenes with highly variable elements, or categories with significant intra-class topological differences would require alternative strategies. In such cases, predefined ROIs might not adequately capture the relevant information, or might even be detrimental by focusing on irrelevant or inconsistent features. Future research could explore more flexible or adaptive mechanisms for defining regions of interest, enabling the application of similar principles to a wider range of object recognition tasks.

While SapiensID achieves state-of-the-art performance in long-term ReID, its short-term ReID accuracy lags behind methods like Soldier~\cite{chen2023beyond} and HAP~\cite{yuan2023hap}. This discrepancy stems from a fundamental conflict between short-term cues—such as clothing—and long-term biometric traits like facial features and body shape. Soldier and HAP leverage masked reconstruction objectives that emphasize visible appearance cues, including clothing, making them more effective for short-term scenarios. In contrast, SapiensID is trained on the WebBody4M dataset, which features frequent clothing changes and thus prioritizes identity over appearance. Addressing this trade-off remains an open challenge, and future work could explore unified models that balance both short-term appearance cues and long-term identity features.

\section{Ethical Concerns}

Our goal is to facilitate research in human recognition while operating strictly within the bounds of copyright law, privacy regulations, and ethical considerations. For large-scale image datasets, it is a common practice to release datasets in URL format~\cite{laion, beaumont-2021-img2dataset} because researchers do not hold the rights to redistribute the data directly. 
By providing permanent link URLs, labels and a one step code to download and prepare dataset, researchers can have access and utilize the data responsibly, while respecting the rights of copyright holders and individuals. 
We believe this approach balances the need for large-scale datasets to advance research with the imperative to protect intellectual property and privacy.

\end{document}